\documentclass[11pt]{article}

\usepackage{graphicx}
\usepackage{amsmath}
\usepackage{amssymb}
\usepackage[a4paper,left=1.9cm,right=1.9cm,top=1.7cm,bottom=1.7cm]{geometry}
\usepackage[T1]{fontenc}
\usepackage{lmodern}
\usepackage{microtype}
\usepackage{booktabs}
\usepackage{caption}
\usepackage{tocloft}
\usepackage[hidelinks,breaklinks=true]{hyperref}
\usepackage[numbers,sort&compress]{natbib}
\usepackage[capitalize,noabbrev, nameinlink]{cleveref}

\usepackage{enumitem}
\usepackage{array}
\newcolumntype{L}[1]{>{\raggedright\arraybackslash}p{#1}}
\usepackage{xcolor}
\usepackage{float}
\usepackage[ruled,vlined]{algorithm2e}
\usepackage{listings}
\usepackage{xurl}
\usepackage{pifont}
\usepackage{makecell}
\usepackage{tabularx}
\newcommand{\cmark}{\ding{51}} 
\newcommand{\xmark}{\ding{55}} 

\definecolor{biocol}{HTML}{006E6E}

\definecolor{codebg}{HTML}{F8F6F0}
\definecolor{coderule}{HTML}{E3DDCD}
\definecolor{codekey}{HTML}{8A5A00}
\definecolor{codecom}{HTML}{6E7C86}
\newsavebox{\tbdbox}
  {\par\medskip\noindent\begin{lrbox}{\tbdbox}%
   \begin{minipage}{\dimexpr\linewidth-2\fboxsep-2\fboxrule}\small}%
  {\end{minipage}\end{lrbox}\fbox{\usebox{\tbdbox}}\par\medskip}

\usepackage{parskip}

\title{SeqMaestro: From nucleotide sequences to biological hypotheses through interpretable machine learning}

\author{%
Evgeny S. Saveliev$^{1}$, Krzysztof Kacprzyk$^{1}$,\\
Charlotte Capitanchik$^{2,3}$, Neelanjan Mukherjee$^{4}$, Kate Matlin$^{4}$, Ryan Sheridan$^{4}$,\\
Srinivas Ramachandran$^{4}$,
Jernej Ule$^{2,3}$, David L. Bentley$^{4}$, Mihaela van der Schaar$^{2,1}$\\[0.5em]
\small $^{1}$Department of Applied Mathematics and Theoretical Physics, University of Cambridge, Cambridge, UK\\
\small $^{2}$The Francis Crick Institute, London, UK\\
\small $^{3}$UK Dementia Research Institute at King's College London, London, UK\\
\small $^{4}$Department of Biochemistry and Molecular Genetics, University of Colorado Anschutz Medical Campus,\\
\small Aurora, CO, USA\\[0.5em]
}

\date{}

\begin{document}
\maketitle

\begin{abstract}
Nucleotide sequence analysis is central to problems spanning regulatory genomics, evolutionary biology, and phenotype prediction. Classical bioinformatics methods extract interpretable sequence properties such as motifs and k-mer composition, but their flexibility is limited. In contrast, modern deep learning models can learn powerful predictive representations directly from raw sequences, yet their internal representations and decision mechanisms are difficult to inspect. Interpretable machine learning methods (e.g., sparse linear models and decision trees) provide human-understandable representations of predictive relationships but are not designed to operate directly on nucleotide sequences. 
Here, we introduce SeqMaestro, a machine learning framework that proposes biological hypotheses from nucleotide sequences using interpretable models. Our solution is centered around a two-layer interface that connects nucleotide sequences with the broader ecosystem of interpretable machine learning.
SeqMaestro uses this interface to fit diverse combinations of interpretable models, feature representations, and extraction strategies, leveraging variability across transparent models to identify robust biological signals and richer predictive relationships than feature importance alone can provide. The system also supports data transformation and cleaning, model fitting, hyperparameter tuning, reliability analysis, and synthesis of results into a contextualized written report. By providing these capabilities through a no-code workflow, SeqMaestro is designed to make interpretable sequence analysis accessible to researchers without requiring extensive programming or machine learning expertise. SeqMaestro thereby provides an accessible route from nucleotide sequences to biological hypotheses.
\end{abstract}

\section*{Introduction}

Nucleotide sequences encode the information required for life and underlie diverse biological processes in health and disease, including gene regulation and RNA processing, replication, genome maintenance, and mutation. Thus, they can be a great source of novel biological hypotheses and discoveries after a careful computational analysis. Indeed, bioinformatics analyses routinely involve attempts to extract biologically meaningful features from nucleotide sequences, a process that often relies on labour-intensive trial and error. Traditionally, such analysis relied on specialized bioinformatics methods that capture biologically meaningful sequence structure, common approaches including: homology-based search with tools such as BLAST \citep{altschul1990basic}, motif discovery with methods such as MEME \citep{bailey2006meme}, probabilistic sequence models such as hidden Markov models \citep{burge1997prediction}, explicit sequence representations based on k-mer composition \citep{moeckel2024survey}, and supervised predictive methods built on related representations, such as gkm-SVM \citep{ghandi2014enhanced}. These approaches are often highly effective, but they typically require the researcher to specify in advance the form in which a relevant sequence signal is expected to appear. They define a particular vocabulary in which discoveries can be made, such that biologically relevant properties outside that vocabulary may be overlooked.

Deep learning has substantially relaxed this requirement by learning predictive sequence representations directly from nucleotide sequences. Models such as DeepSEA \citep{zhou2015predicting} learn regulatory sequence features using convolutional neural networks, whereas Enformer \citep{avsec2021effective} integrates long-range sequence context using transformer-based architectures. More recent genomic foundation models, such as Nucleotide Transformer \citep{dalla2025nucleotide}, extend this paradigm by learning general-purpose sequence representations through large-scale pretraining. These approaches have substantially expanded the range and complexity of biological phenotypes and molecular events that can be predicted from sequence. Yet prediction alone is often insufficient for scientific discovery: a highly accurate black-box model may reveal the existence of a signal without explaining which sequence properties drive it or how those properties interact. Post-hoc explanation methods can highlight sequence positions or patterns associated with model predictions, but they do not generally yield a predictive model whose logic is itself expressed in explicit rules. Moreover, post-hoc explanations may not be reliable and can be misleading \citep{rudin2019interpretable}. Some explanation methods may not faithfully reflect the information learned by the underlying model \citep{adebayo2018sanity}; explanations can be unstable under small input perturbations \citep{ghorbani2019interpretation}, and different methods can produce inconsistent explanations of the same prediction \citep{chen2024applying}. Rather than explaining a complex predictive model after training, we therefore ask whether the predictive model itself can serve as an explicit biological hypothesis.

\begin{figure}[h!]
    \centering
    \includegraphics[width=0.9\linewidth,trim={0.2cm 0cm 0.2cm 0},clip]{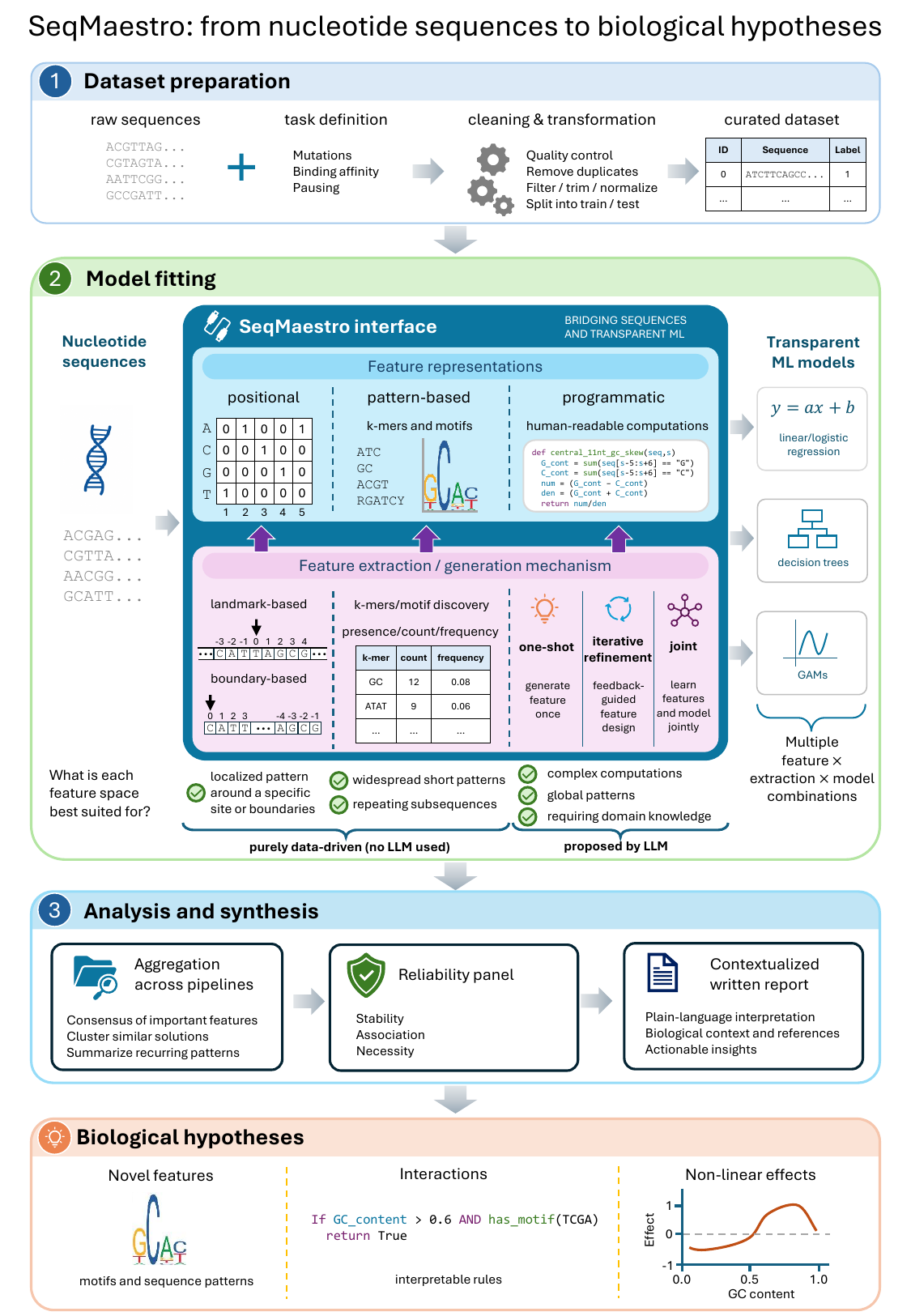}
    \caption{In this paper, we build a bridge between nucleotide sequences and interpretable machine learning.}
    \label{fig:big_figure}
\end{figure}

Intrinsically interpretable (or transparent) machine learning constructs models whose predictive mechanisms can be inspected directly. That includes linear models, decision trees, generalized additive models (GAMs), and rule-based methods. However, this methodological ecosystem remains largely disconnected from nucleotide sequence analysis. Most interpretable learning algorithms operate on well-defined prediction tasks with specified targets and structured tabular features rather than raw sequences. 

To address this challenge, we construct an interface between nucleotide sequences and transparent machine learning that maps raw sequences into a broad, searchable space of human-readable hypotheses, allowing us to utilize the existing ecosystem of interpretable machine learning models for sequence-based prediction and discovery. The interface has two layers. First, we introduce a taxonomy of interpretable sequence representations that separates three levels of feature expressivity: positional features, pattern-based features such as k-mers and motifs, and programmatic features defined by explicit, human-readable computations over a sequence. These representations form increasingly expressive languages for describing sequence-level hypotheses.
Because intrinsically interpretable models can incorporate only a limited number of features and interactions while remaining understandable, richer biological structure must often be captured within the features themselves. Programmatic features provide a particularly flexible representation, allowing composition, positional relationships, motif organization, and higher-order logical structure to be expressed while remaining directly inspectable. This expands the space of interpretable hypotheses that can be explored without requiring the researcher to specify in advance the particular sequence properties that are likely to be relevant.

The second layer is the actual mechanism of extracting these features. In particular, it concerns three complementary strategies for programmatic feature generation.
These strategies differ in how tightly hypothesis generation is coupled to empirical feedback: from one-shot proposal, through iterative propose-fit-refine cycles, to joint procedures in which feature construction becomes part of model fitting itself. This interface allows us to fit different combinations of transparent ML models, feature types, and even extraction strategies. 

Here we introduce this interface within a framework called SeqMaestro. SeqMaestro is an end-to-end framework for searching nucleotide sequences for explicit, predictive, and testable biological hypotheses. It formalizes biological questions as supervised sequence-based prediction tasks and then leverages the Rashomon Effect by fitting multiple combinations of transparent model pipelines as defined by the interface. Different transparent model classes expose different forms of biological structure. Linear models identify additive associations, GAMs reveal nonlinear response relationships, and decision trees expose thresholds and conditional dependencies between sequence properties. SeqMaestro therefore treats the model class itself as part of the hypothesis search, fitting diverse, transparent pipelines, aggregating their results, and assessing the robustness of the resulting discoveries.

Crucially, SeqMaestro does not restrict interpretation to individual feature importance. Because the fitted models are themselves transparent, the resulting hypothesis may be an individual feature, a nonlinear response curve, a threshold, or a conditional interaction between sequence properties. This allows us to move from identifying isolated signals (e.g., that CpG frequency is predictive) to recovering richer predictive logic. A decision tree, for instance, may reveal that a particular sequence property is informative only when another condition is satisfied (i.e. only in a particular sequence context), whereas a GAM can show how the prediction changes continuously across the values of a feature, such as GC content. In this sense, the predictive model itself becomes part of the biological hypothesis.

SeqMaestro also performs the practical steps required for end-to-end analysis, including data transformation and cleaning, model fitting, hyperparameter tuning, reliability analysis, and synthesis of results into a contextualized written report. 
This automation makes it practical to explore a hypothesis space that would otherwise require substantial manual feature engineering, model comparison, and interpretation.
By providing these capabilities through a no-code workflow, SeqMaestro is designed to make interpretable sequence analysis accessible to researchers without requiring extensive programming or machine learning expertise. By jointly treating feature discovery, model fitting, and biological interpretation as components of a single workflow, SeqMaestro provides a general framework for moving from nucleotide sequences to explicit, predictive, and testable biological hypotheses.

\section*{Results}

\subsection*{SeqMaestro methodology}

SeqMaestro framework has three stages: (1) dataset preparation, (2) model fitting, and (3) analysis and synthesis. They are depicted in \cref{fig:big_figure}.

\subsubsection*{Dataset preparation}

The goal of the first stage is to prepare a curated dataset that can be passed to our interface in the second stage. In particular, it transforms the raw sequences into a form suitable for \textbf{nucleotide sequence-based prediction}, in which each observation consists of a nucleotide sequence $x_i$ and a binary label $y_i$. Our current implementation focuses on binary classification, while regression and multiclass or multilabel prediction represent natural extensions.

Many biological questions can be formulated in this way. For example, in RNA polymerase II pausing, fixed-length sequence windows around genomic positions can be labeled according to whether strong pausing occurs; similarly, in somatic hypermutation, local sequence windows can be labeled according to whether the central nucleotide is mutated, enabling the discovery of sequence features associated with increased mutation probability. This stage also includes quality controls and splitting the sequences into train and test (holdout) samples. Holdout samples are not used for model fitting and are only used for evaluation.

\subsubsection*{Model fitting - SeqMaestro interface}

SeqMaestro fits multiple intrinsically interpretable models, including logistic regression, additive models, and decision trees, across the different feature representations and extraction mechanisms described below. We refer to each such pipeline as a separate \textit{arm} and to the whole collection of arms as a \textit{battery}. Together, these approaches allow us to compare not only different interpretable model classes and feature representations, but also different strategies for searching the space of interpretable programmatic features.

Nucleotide sequences are naturally represented as strings over a small alphabet, typically $A$, $C$, $G$, and $T$ for DNA or $A$, $C$, $G$, and $U$ for RNA. Although this raw representation is biologically natural, most conventional interpretable machine learning methods are not designed to operate directly on nucleotide strings. Applying such models therefore requires transforming each sequence into a structured feature space.

We consider three broad families of feature representations: positional features, pattern-based features, and programmatic features.

\paragraph{Positional features.}

The most direct representation treats each sequence position as an individual feature. For a sequence of length $L$, this produces $L$ categorical variables, where the feature at position $i$ corresponds to the nucleotide observed at that position.

To accommodate sequences of varying lengths, we introduce two encoding variants: landmark-based and boundary-based. Both of them operate with a predefined window (e.g., 50 nucleotides). The landmark-based system indexes positions from a particular point of a sequence in both directions. Boundary-based, on the other hand, focuses only on nucleotides at the beginning and end of the sequences. The choice of encoding strategy can be specified by the user or suggested by SeqMaestro itself, based on background knowledge of the problem. 

This representation preserves a part of the positional structure of the original sequence while expressing it in a form that can be consumed by conventional machine learning algorithms. Models that do not natively support categorical variables require an additional encoding step. For example, one-hot encoding can represent each nucleotide at each position with binary indicator variables, enabling models such as logistic regression to use it.

\paragraph{Pattern-based features.}

A second family of representations describes sequences through the occurrence or strength of local sequence patterns. This category includes both exact subsequences, such as k-mers, and more flexible patterns, such as motifs.

For k-mer representations, features can describe the presence, count, or frequency of specific nucleotide subsequences of length $k$. For example, a feature may indicate whether a particular 6-mer occurs in a sequence or how many times it appears. Multiple values of $k$ can be considered simultaneously, allowing patterns at different local scales to be represented.

Motif-based representations generalize this idea by allowing patterns to tolerate variation across positions. A motif may be represented, for example, by a consensus sequence or a position weight matrix. The motif representation itself is not necessarily passed directly to the predictive model. Instead, it defines a matching procedure from which conventional scalar features can be derived, such as the maximum motif-match score within a sequence, the presence of a match above a threshold, or the number of motif occurrences. The motifs themselves can be suggested by standard motif discovery algorithms.

Thus, both k-mers and motifs act as pattern detectors that transform raw sequences into Boolean, integer, or continuous variables suitable for standard interpretable machine learning methods. The principal distinction is that k-mers generally define exact sequence matches, whereas motifs describe generalized or probabilistic local patterns.

\paragraph{Programmatic features.}

A substantially more expressive feature space can be obtained by allowing features to be defined as explicit, interpretable computations over the sequence. Such features may represent global compositional properties; positional conditions; counts and distances between sequence elements; motif or k-mer relationships; palindromic or symmetric structure; or logical combinations of multiple conditions. For example, a feature might indicate whether a compositional statistic exceeds a threshold while a particular nucleotide occurs at a specified position, or whether two motifs occur within a given distance of one another.

Programmatic features therefore subsume many simpler feature types as possible building blocks. Positions, k-mers, motifs, sequence-composition statistics, and other primitive operations can be composed into higher-level, human-readable hypotheses about sequence structure.

The hypothesis space of possible programmatic features is extremely large, making exhaustive enumeration impractical. In our framework, candidate features are instead proposed by a feature-generation procedure powered by large language models. We consider three distinct strategies for feature generation. In the \textbf{one-shot setting}, a language model proposes a set of candidate features once, which are subsequently evaluated by the interpretable model. In the \textbf{iterative setting}, features are proposed and evaluated over multiple rounds: after each model fit, information about the resulting predictive performance and selected features is returned to the language model, which uses this feedback to propose improved features for the next iteration. Finally, in the \textbf{joint setting}, feature generation is embedded directly into the model-fitting procedure, so that new programmatic features are constructed adaptively as the model is learned. In the current implementation, this joint strategy is available for decision trees through an existing method \citep{huynh2026deft}.

Conceptually, the three representations form a progression from direct sequence encoding to increasingly abstract descriptions of sequence properties: positional features describe individual locations, pattern-based features describe recurring local sequence patterns, and programmatic features represent general interpretable functions of the sequence. The main drawback of lower-complexity features is their limited expressivity within a fixed interpretability budget. For instance, if we want to use a decision tree with depth 3, sole dependence on positional features will make it harder to discover complex patterns. However, as our case studies below show, all feature types have their advantages. Positional features are best suited for capturing short, local patterns around a specific site or its boundaries. K-mers and motifs excel at identifying short, widespread, or recurring patterns. Programmatic features are ideal for situations involving global patterns, where we need to perform complex computations or rely on prior knowledge of the language model (LLM). These features also allow for optional scientist guidance regarding the preferable form of the discovered features. For instance, if the model should focus on global patterns, symmetries or counts. Appendix~\ref{app:features} follows features of every kind, whether derived from the data alone or proposed by the language model under each of these strategies, through the rest of the framework.

\subsubsection*{Analysis and synthesis}

The third stage is divided into two sub-stages. First, the models are reduced to the 10 most important features to ensure that they are understandable. After the features are selected, the reduced models are refitted to the data. The second sub-stage is the reliability panel, which uses various techniques (grouped into three tiers) to assess the robustness of the discovered features. Tier 1 checks the stability of the discovered features by testing whether the same features would appear under a random sub-sampling of the dataset. Tier 2 assesses the association between the target and each feature separately. Finally, Tier 3 assesses necessity. It checks whether the model loses accuracy when a particular feature is removed. After the reliability panel is completed, SeqMaestro prepares a written report on the entire analysis.

\subsection*{Experimental Findings}

SeqMaestro aims to give a biologist two things from a set of labeled sequences: interpretable model(s) small enough that their internal logic can be understood, and a tractable list of sequence properties that are dependable enough to follow up as hypotheses in subsequent experiments or analyses.
To see whether it accomplishes these goals, we applied SeqMaestro to four biological questions, two about DNA and two on RNA, chosen to cover a range of different informative signals that the sequences contain.

Each question is discussed in turn below and presented in individual figures.
The four figures share a common layout for ease of comparison: the task and its window (panel \textbf{a}), how every configuration fares when it is cut down to ten features (panel \textbf{b}), one interpretable model presented in full (panels \textbf{c} and \textbf{d}), and the concepts that several configurations converged on, with the evidence behind each (panel \textbf{e}).
Where available, we show independent or orthogonal support for novel features identified in SeqMaestro analysis, in panel \textbf{f}.
Gradient-boosted trees serve primarily as a black-box comparison and do not meaningfully surpass interpretable models on any task.
We use AUROC as the primary performance metric throughout the report and figures.

\subsubsection*{Polycomb nucleation}

\begin{figure}[p]
\centering
\includegraphics[width=0.98\linewidth]{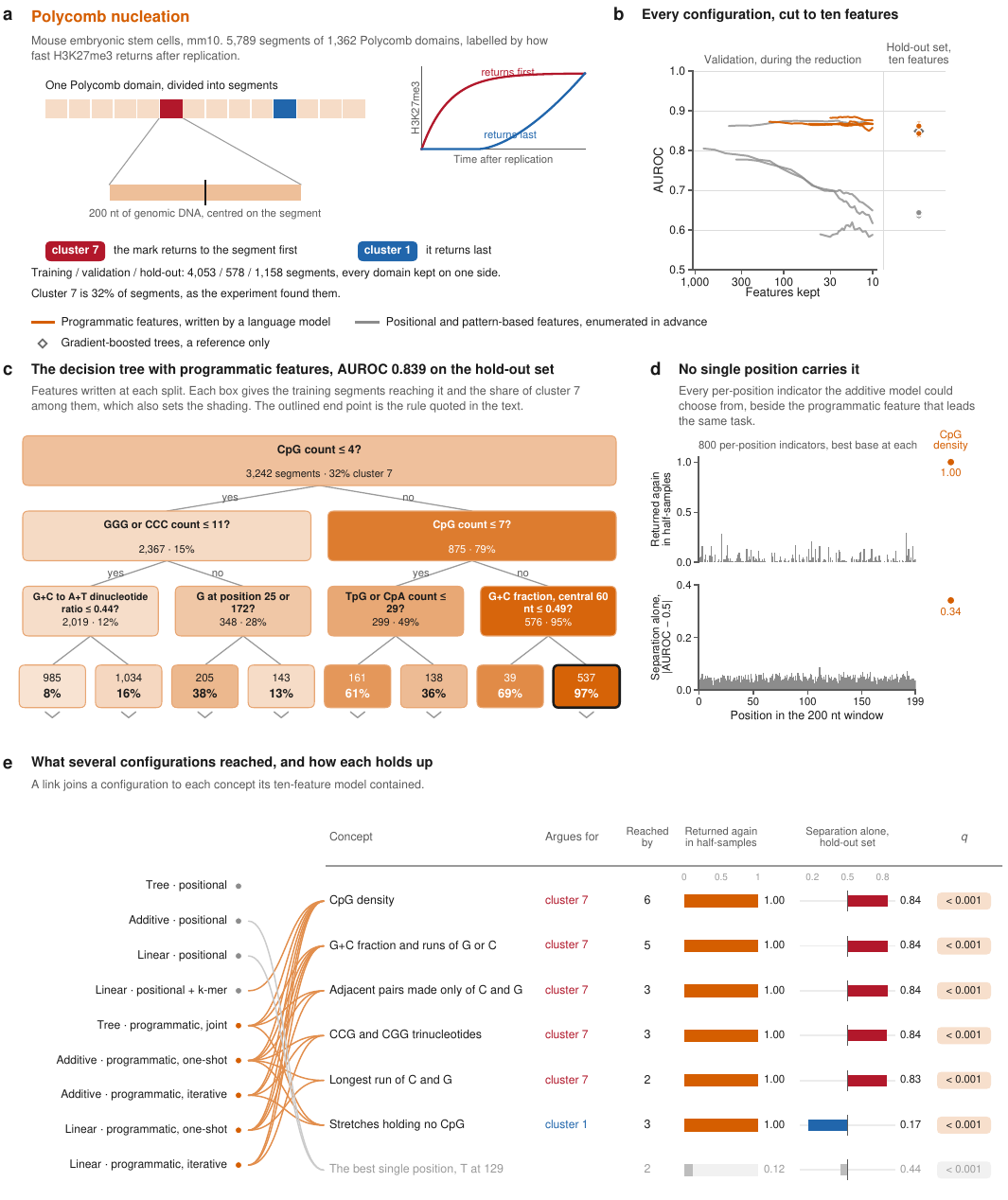}
\caption{\textbf{Polycomb nucleation.}
\textbf{a}, The task. Segments of mouse Polycomb domains are labeled by how fast H3K27me3 returns after replication, the fastest-recovering cluster of segments (cluster 7) against the slowest (cluster 1), and each model sees 200 nt of genomic DNA centered on the segment.
\textbf{b}, Every interpretable configuration reduced from all the features it selected down to ten. Curves are validation AUROC recorded during the reduction. Dots at the right are hold-out AUROC of the ten-feature model, and the open diamond is the best gradient-boosted configuration at ten features.
\textbf{c}, The decision tree with programmatic features written at each split. Boxes give the training segments reaching them and the share of cluster 7 among those segments, also reflected in the shading.
\textbf{d}, Every per-position indicator the additive model could choose from, laid along the window: the proportion of 100 half-samples that returned it (measure of robustness), and how far it separates the hold-out segments as a score on its own, beside both numbers for CpG density.
\textbf{e}, Concepts reached by more than one configuration. A link joins each configuration to the concepts among its ten features. Each row gives how many configurations reached the concept, the
proportion of half-samples that returned (measure of robustness), how far it separates the hold-out segments on its own, and the $q$-value of that separation after correction for the whole pool. A grey row fails at least one tier.}
\label{fig:r1}
\end{figure}

Polycomb domains are stretches of chromatin carrying the repressive histone mark H3K27me3.
Replication halves this mark, and it is restored over the following cell cycle, first at a small number of nucleation sites and then by spreading outward from them.
\citet{veronezi2024nucleation} followed that recovery across 1,362 domains in mouse embryonic stem cells, dividing each into segments and clustering the segments by their recovery. We took the fastest-recovering segments, their cluster 7, as one class and the slowest, their cluster 1, as the other, and gave every model 200 nt of genomic DNA centered on the segment (Fig.~\ref{fig:r1}a).
Because both classes come from the same domains, the comparison is between segments that share the same chromatin context, and it asks what distinguishes the places where the mark returns first versus last.

SeqMaestro identifies one concept as especially important for this problem. Six of the nine interpretable configurations placed a measure of CpG density among their ten features: the decision tree with joint programmatic features, both additive and both sparse linear models with programmatic features, and the sparse linear model with positional and k-mer features. Only the three configurations restricted to per-position indicators, which cannot express a count, did not. Between them the six wrote the same quantity in eight different forms, from a plain CpG count and a CpG density to the observed-over-expected CpG ratio and the number of distinct CpG-containing 6-mers, which the reliability panel grouped into one concept because they compute near-identical values on the same sequences, and every one of those forms proved reliable in our robustness analysis. Moreover, CpG density also separates the hold-out segments at 0.84 AUROC on its own (Fig.~\ref{fig:r1}e).
SeqMaestro builds on this and identifies (using a decision tree) a simple condition that enables us to locate cluster 7 segments with very high precision. If there are more than seven CpG dinucleotides in the 200 bases and the fraction of G and C together in the central 60 bases is above 49\% then 97\% of training segments belong to cluster 7 (Fig.~\ref{fig:r1}c).
The reduction to ten features then allows us to check whether the signal is spread out over the 200-base window or localized at a specific position. After the reduction, programmatic features hold their score while models depending only on positional features fall from about 0.81 to about 0.64 (Fig.~\ref{fig:r1}b), and none of the per-position indicators is returned reliably in our robustness analysis (Fig.~\ref{fig:r1}d).
Thus, what marks a cluster 7 segment within these 200 bases is spread across the window, with no particular coordinate highly influencing it.

The association between Polycomb recruitment and unmethylated CpG islands is among the best established in chromatin biology.
KDM2B binds unmethylated CpG and recruits a variant PRC1 complex, whose H2A ubiquitylation in turn recruits PRC2 \citep{farcas2012kdm2b, blackledge2014variant}, and GC-rich elements are sufficient to nucleate H3K27me3 where they are inserted \citep{mendenhall2010gcrich}.
Recovering that from sequence alone, with no chromatin measurement of any kind, is the starting point, and it is achieved by SeqMaestro.
Two aspects of SeqMaestro analysis go beyond this.
The comparison here is within domains, so the usual statement that Polycomb targets CpG islands is already true of every segment, and the result says that CpG density also decides where inside a domain the mark is re-established first after replication \citep{reveron2018accurate, oksuz2018capturing}.
That connects the recruitment which establishes a domain to the way the domain is maintained through cell division, and it can be checked directly against maps of nucleation sites defined by PRC2 activity in a follow-up investigation.
The second is the shape of the effect. The tree's rule describes a threshold, and the curve the additive model fits to CpG density rises steeply and then flattens (Appendix~\ref{app:reports}).
This leads to the hypothesis that Polycomb nucleation responds to CpG density as a switch rather than a gradient. This could be tested by inserting a series of synthetic elements with CpG density spanning the fitted threshold and measuring whether re-establishment turns on sharply or increases gradually.

\subsubsection*{RNA polymerase II pausing}

\begin{figure}[p]
\centering
\includegraphics[width=0.98\linewidth]{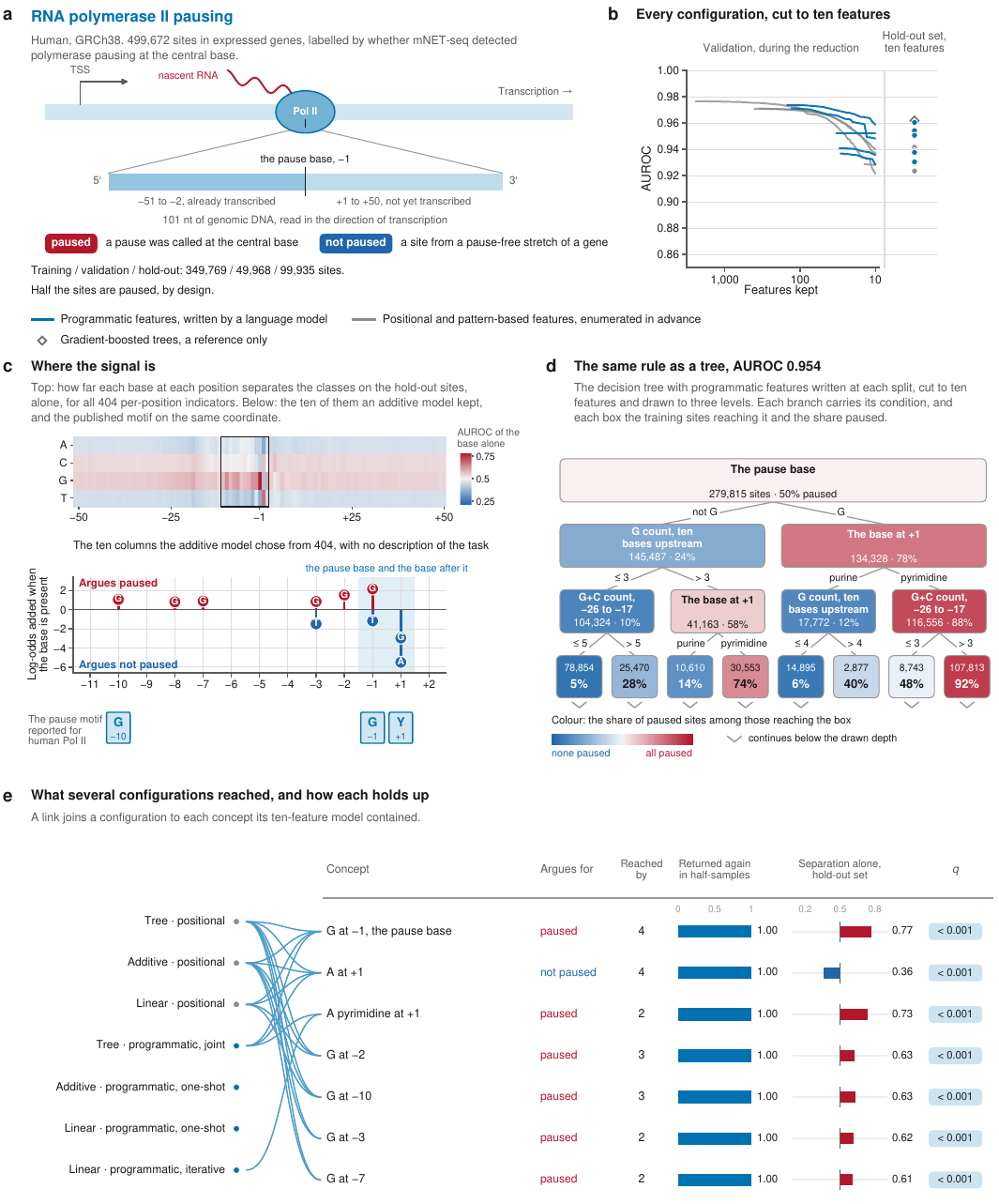}
\caption{\textbf{RNA polymerase II pausing.}
\textbf{a}, The task. Sites in expressed human genes are labeled by whether mNET-seq detected a pause at the central base, and each model sees a 101 nt read in the direction of transcription, with the pause base, the 3$'$ end of the nascent RNA, at position $-1$ and the base after it at $+1$, as the published pause motifs are numbered. The window runs from $-51$ to $+50$.
\textbf{b}, Every interpretable configuration reduced to ten features, drawn as in Fig.~\ref{fig:r1}b.
\textbf{c}, Top: how far each base at each position separates the classes on the hold-out sites as a score on its own, for all 404 per-position indicators. Middle: the ten indicators the additive model kept, at the positions they name, with the log-odds the model adds when that base is present. Bottom: the pause motif reported for human polymerase, on the same coordinate, which the model matches at all three positions.
\textbf{d}, The decision tree with programmatic features written at each split, cut to ten features and drawn to three levels of questions. Each branch carries the condition that leads down it. Each box gives the training sites reaching it and the share of them that are paused, which also sets its colour, and a chevron marks a branch that continues below the drawn depth.
\textbf{e}, Concepts reached by more than one configuration, drawn as in Fig.~\ref{fig:r1}e.}
\label{fig:r2}
\end{figure}

RNA polymerase II does not transcribe at a constant rate. It pauses at particular bases, and mNET-seq \citep{nojima2015netseq} detects those pauses as peaks of polymerase density.
We took 499,672 sites in expressed human genes, half of them pauses and half drawn from pause-free stretches of the same genes, and gave every model 101 nt of genomic DNA read in the direction of transcription, with the pause base at position $-1$ and the base after it at $+1$ (Fig.~\ref{fig:r2}a).

The SeqMaestro feature-model battery shows that, in this task, several informative signals lie at fixed distances from the pause.
Every interpretable configuration with ten features achieves AUROC between 0.92 and 0.96 (Fig.~\ref{fig:r2}b).
The additive model (GAM) with positional features yielded the clearest insights. Initially, it had 404 per-position indicators and no description of the task. After reduction, all ten features were within positions $-10$ to $+1$ (Fig.~\ref{fig:r2}c).
In particular, G at the pause base argues for a pause, A at $+1$ argues strongly against, and G at each of five positions upstream, $-2$, $-3$, $-7$, $-8$ and $-10$, argues for a pause.

Several of these features agree with the consensus motifs reported for human pause sites \citep{gajos2021conserved, sheridan2019widespread} \citep{Fong2022}.
Most notably, the G at $-10$ is conserved in the consensus pause element of \textit{E.~coli} \citep{larson2014pause, vvedenskaya2014interactions}.
The $-10$ position corresponds to the first base of the transcription bubble, and translocation requires breaking the rNdN base pair at this position.
rGdC is a particularly stable base pair, whose disruption hampers translocation when it is situated at $-10$.
The identification of G at $-10$ by SeqMaestro (Fig.~\ref{fig:r2}c) as a significant feature of RNA polymerase II pause sites strongly suggests that the same mechanism of inhibiting translocation contributes to pausing by both the human and the \textit{E.~coli} enzymes.
The possible functional significance of G at $-2$, $-3$, $-7$ and $-8$ in promoting pausing (Fig.~\ref{fig:r2}c) may merit future mechanistic investigation.

The decision tree with programmatic features (Fig.~\ref{fig:r2}d) asks first whether the pause base is G and whether the base at $+1$ is a pyrimidine, and then recovers a similar enrichment of G over the ten bases upstream of the pause, within the transcription bubble, as a feature of pause sites.
In addition, it identified G+C content at positions $-26$ to $-17$ as a further predictor of pausing.
These positions correspond to the first bases of the RNA to emerge from the exit channel of the polymerase, and one hypothesis is that secondary structure in the emerging transcript contributes to pausing \cite{Kang2018}.
Every one of these properties passed our robustness checks and holds on the hold-out sites, and those reached by more than one configuration are listed in Fig.~\ref{fig:r2}e.

\subsubsection*{eIF4E dependence}

\begin{figure}[p]
\centering
\includegraphics[width=0.98\linewidth]{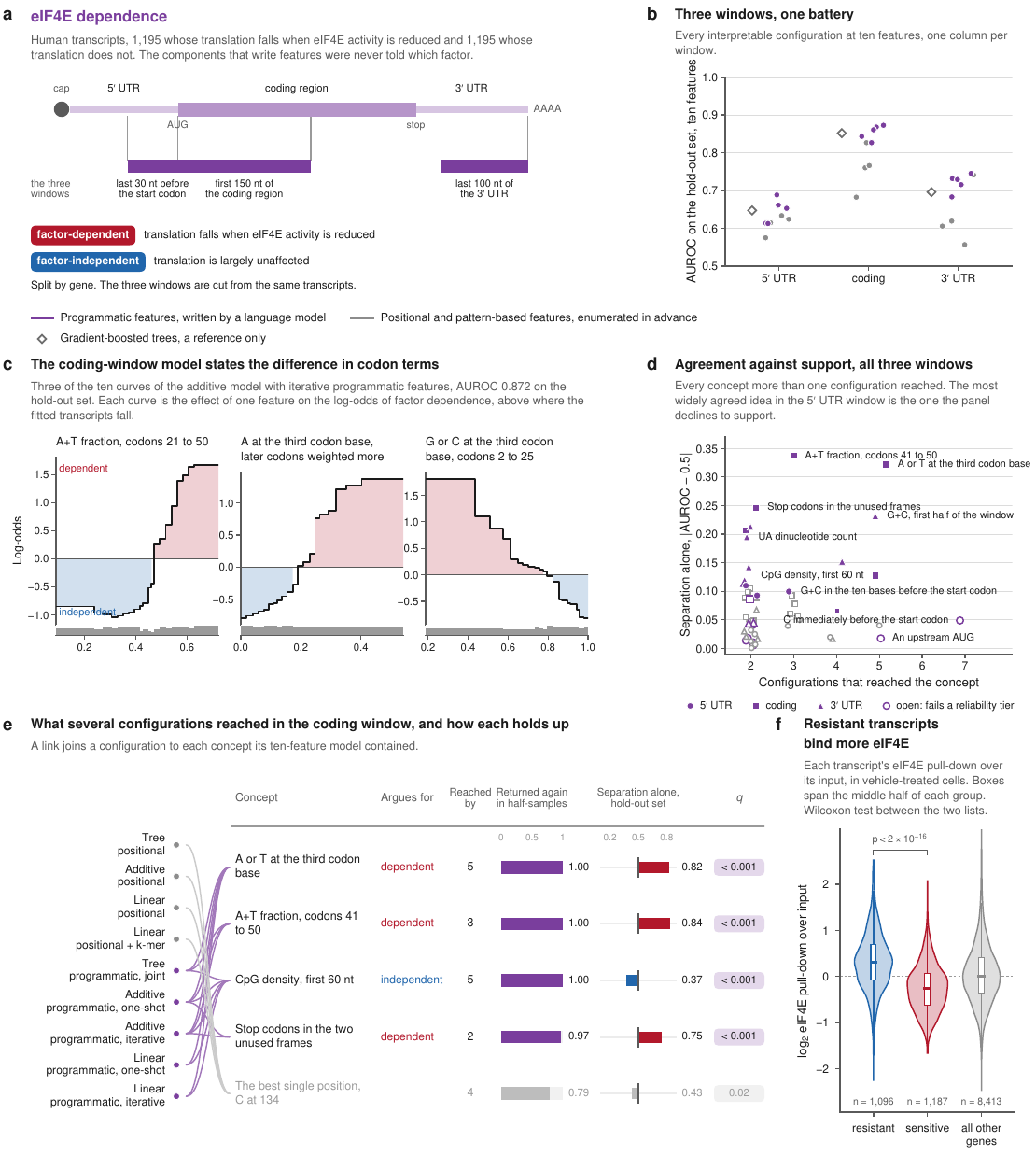}
\caption{\textbf{eIF4E dependence, three windows of one transcript set.}
\textbf{a}, The task. Two lists of human transcripts, one whose translation falls when eIF4E activity is reduced and one whose translation does not, windowed three ways.
\textbf{b}, Every interpretable configuration at ten features, one column per window. Marks as in Fig.~\ref{fig:r1}b.
\textbf{c}, Three of the ten curves of the additive model fitted to the coding window. Each is the effect of one feature on the log-odds of factor dependence, above a strip showing where the fitted
transcripts fall.
\textbf{d}, Every concept that more than one configuration reached, in all three windows, placed by how many configurations reached it and how far it separates the hold-out transcripts on its own. Open
marks fail at least one tier of the reliability panel.
\textbf{e}, Concepts reached by more than one configuration in the coding window, drawn as in Fig.~\ref{fig:r1}e.
\textbf{f}, How much eIF4E each transcript binds, measured as its enrichment in an eIF4E immunoprecipitation over the input in vehicle-treated cells, for the eIF4E-sensitive list (the factor-dependent class of \textbf{a} to \textbf{e}), the eIF4E-resistant list (factor-independent) and all other genes, coloured as the classes in \textbf{a}. Boxes span the middle half of each group, with the median across them. The two lists differ by a Wilcoxon test, $p < 2 \times 10^{-16}$. Reanalysed from the immunoprecipitation data of \citet{Roiuk2024}.}
\label{fig:r3}
\end{figure}

The translation initiation factor eIF4E, the cap-binding component of the eIF4F complex, plays a central role in cap-dependent translation and is regulated by nutrient-sensing pathways. In previous work, ribosome profiling was used to identify mRNAs that remain efficiently translated when eIF4E is inhibited (eIF4E-resistant) versus those whose translation is blocked (eIF4E-sensitive) \cite{Roiuk2024}. From this study, we generated two lists of the most eIF4E-resistant and eIF4E-sensitive transcripts, totaling 1,195 transcripts each. Because transcript lengths differ between these groups, and mRNA regions (5$'$ UTR, CDS, 3$'$ UTR) have different regulatory functions \cite{Mignone2002}, evolutionary constraints \cite{Shabalina2004}, and relative length and nucleotide composition \cite{Pesole2001}, we focused on three local windows: the thirty nucleotides before the start codon, the first 150 nucleotides of the coding region, and the last 100 nucleotides of the 3$'$ UTR (Fig.~\ref{fig:r3}a). The LLM-driven feature generation was not told which factor was involved, precluding any recall of literature on eIF4E.

The Kozak sequence, which comprises the ten nucleotides flanking the AUG start codon, is a well-established determinant of translation initiation efficiency, and thus an expected discriminating feature. The canonical Kozak consensus (GCCRCCAUGG) is GC-rich and promotes strong cap-dependent translation. Transcripts with strong Kozak sequences are known to be more resistant to eIF4E inhibition \cite{Acevedo2018}. SeqMaestro identified elevated G and C content in positions 20 to 29 (immediately upstream of the start codon) as a discriminating feature, recovering this expected relationship and validating that the approach identifies functionally relevant sequence determinants.

The most striking finding is that features in the coding sequence, historically associated with translation elongation, emerged as the strongest predictors of eIF4E sensitivity (Fig.~\ref{fig:r3}b). eIF4E-resistant mRNAs were enriched in codons with G or C at the third position (GC3, which Fig.~\ref{fig:r3}e lists in its complementary form, A or T at the third codon base). GC3 content separated hold-out transcripts at AUROC 0.82. GC3 defines optimal codon usage in human cells and is associated with enhanced mRNA stability and translation efficiency \cite{Hia2019}. Recent work showed that mRNAs bearing optimal codons preferentially bind the initiation factors eIF4E and eIF4G1 \cite{Barrington2023}. If codon optimality determines eIF4E resistance through enhanced initiation factor recruitment, we would predict that eIF4E-resistant mRNAs should preferentially associate with eIF4E at baseline. Reanalysis of eIF4E RNA-immunoprecipitation data \cite{Roiuk2024} demonstrated that eIF4E-resistant mRNAs exhibited significantly higher eIF4E binding than sensitive mRNAs and the transcriptome as a whole (Fig.~\ref{fig:r3}f). That SeqMaestro identified GC3 codons as the primary discriminator provides independent support for a mechanism coupling translation elongation kinetics to initiation factor engagement, suggesting that the sequence features that regulate elongation can reciprocally influence how the ribosome recruitment machinery associates with mRNA.

\subsubsection*{PTBP1 exon regulation}

\begin{figure}[p]
\centering
\includegraphics[width=0.98\linewidth]{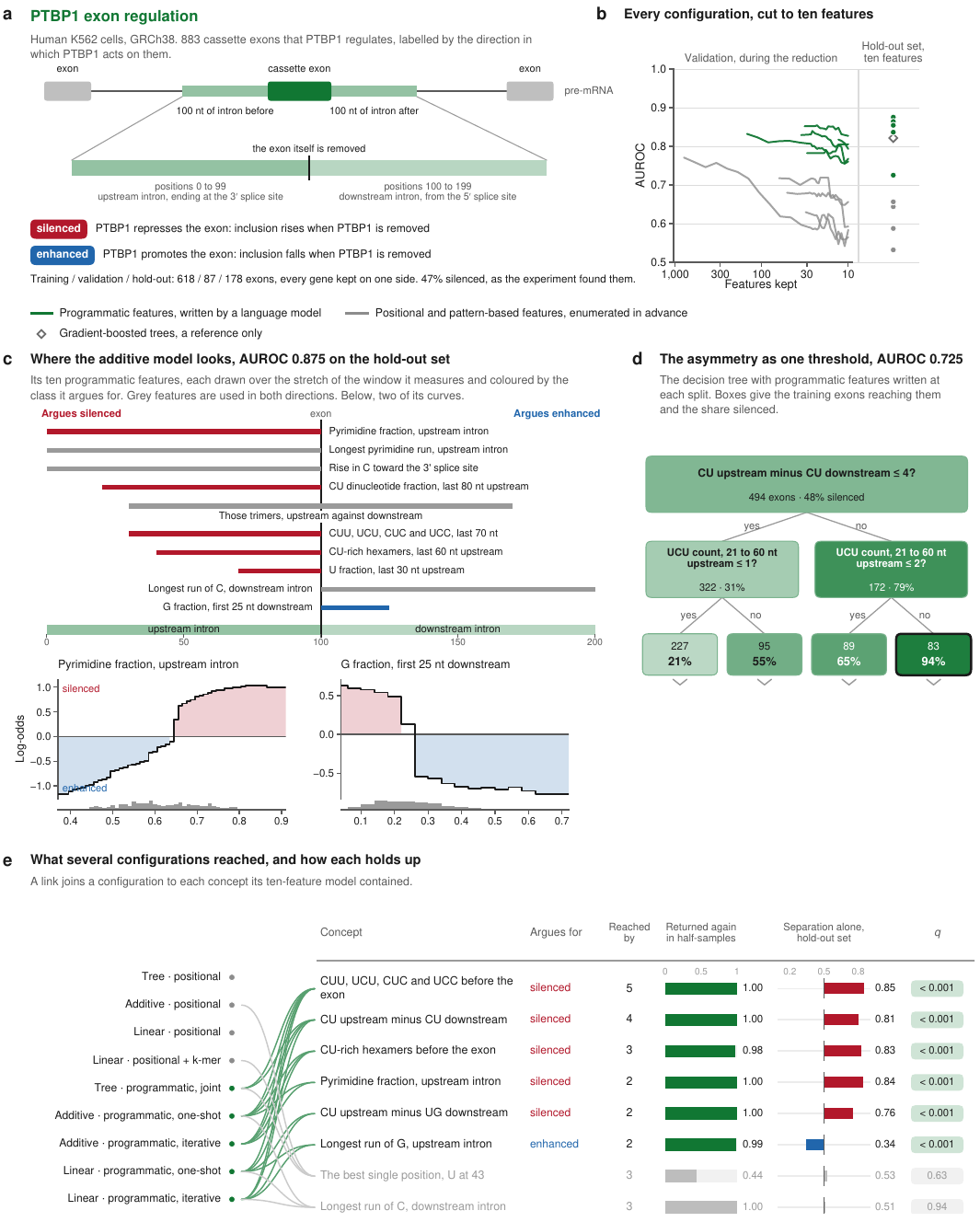}
\caption{\textbf{PTBP1 exon regulation.}
\textbf{a}, The task. Cassette exons that PTBP1 regulates are labeled by the direction of its effect, and each model sees 100 nt of intron on either side of the exon, joined, with the exon itself removed.
\textbf{b}, Every interpretable configuration reduced to ten features, drawn as in Fig.~\ref{fig:r1}b.
\textbf{c}, The ten programmatic features of the additive model, each drawn over the stretch of the window it measures and coloured by the class it argues for, with two of its curves beneath.
\textbf{d}, The decision tree with programmatic features written at each split, to two levels.
\textbf{e}, Concepts reached by more than one configuration, drawn as in Fig.~\ref{fig:r1}e.}
\label{fig:r4}
\end{figure}

A cassette exon is one that is spliced into the mature messenger RNA in some transcripts and skipped in others, thereby usually modulating the abundance or sequence of the resulting protein.
The balance of transcript isoforms is regulated by RNA-binding proteins (RBPs), which can act to enhance or repress splicing based on the positional relationships between their binding sites and exons \citep{Ule2019}.
PTBP1 is an important factor in stem cell self-renewal and cell fate determination, where it is well-characterized as a master regulator of neuronal development \citep{Hu2018}.
It is also broadly expressed across many adult tissues, with roles in cancer, metabolism and immune responses.
PTBP1 is known to repress splicing by binding to pyrimidine-rich sequence elements upstream of the 3$'$ splice site (3$'$SS) and enhance exon inclusion when binding such elements downstream of the 5$'$ splice site (5$'$SS) \citep{Hamid2017, llorian2010position, xue2009genome}.
As the combination of motifs and position encodes the regulation, we were curious if SeqMaestro could learn this regulation \emph{de novo}.
883 cassette exons were annotated as either silenced by PTBP1 (420), or enhanced by PTBP1 (463) \citep{gueroussov2015alternative}.
SeqMaestro was challenged to classify exons as either enhanced or silenced by PTBP1 based on 100 nt upstream and downstream flanking intron sequence alone (Fig.~\ref{fig:r4}a).

This is the task with the widest gap between the two kinds of feature sources (positional and pattern-based versus programmatic).
The best ten-feature model on programmatic features reaches AUROC of 0.875, and the best on positional and pattern-based features 0.656 (Fig.~\ref{fig:r4}b).
The signal is regional pyrimidine content weighted toward the exon, which is challenging to express with ten positional features and short motifs. However, programmatic features can capture it.
The additive model lays its ten features out on the two introns as a map (see Fig.~\ref{fig:r4}c).
Pyrimidine content upstream of the exon, measured at five nested scales (the pyrimidine fraction of the whole 100 nt, the CU dinucleotide fraction of the last 80 nt, the count of the PTBP1-like trimers CUU/UCU/CUC/UCC in the last 70 nt, CU-rich hexamers in the last 60 nt, and the U fraction of the last 30 nt), argues in every case that PTBP1 silences the exon, and G content in the first 25 nt downstream argues that it enhances it.
The decision tree reduces that map to one question: whether there are more than four more CU dinucleotides upstream than downstream, which on its own takes the silenced ratio from 48\% to 79\% (Fig.~\ref{fig:r4}d).
This captures what we know about PTBP1 CU-motif binding being repressive at the 3$'$SS, whilst promoting inclusion at the 5$'$SS, but in a simple, quantitative way that could be applied to sequence design.
Three properties clear every tier of the reliability panel (Fig.~\ref{fig:r4}e): (i) the count of the PTBP1-like pyrimidine trimers CUU, UCU, CUC and UCC in the 70 nt before the exon, reached by five configurations across all three model families; (ii) the difference between CU content upstream and downstream of the exon, reached by four; and (iii) the count of CU-rich hexamers in the 60 nt before the exon, reached by three.
They also separate the hold-out exons at 0.81 to 0.85 on their own.

In conclusion, SeqMaestro is able to condense our knowledge of PTBP1 binding principles into a simple series of quantitative rules, derived from raw sequence \emph{de novo}.
Further, the model extracts several novel features that warrant further investigation.
Firstly, that G-richness in the first 25 nt downstream of the 5$'$SS is predictive of PTBP1 enhanced exons.
One hypothesis is that this feature could represent stabilization of weak 5$'$SS by binding of RBPs such as hnRNPH/F in the presence of PTBP1.
It has been previously shown that PTBP1 can recruit hnRNPH to G-rich sequences in 3$'$ UTRs to regulate polyadenylation \citep{Millevoi2009}.
Secondly, that CU upstream minus UG downstream is predictive of silencing, suggesting that RBPs binding to downstream UG repeats could negate the impact of PTBP1 binding at upstream CU-rich stretches.

\medskip
Taken together, these four tasks demonstrate SeqMaestro's ability to deliver interpretable models and to discover robust sequence properties that predict the target.
On each, it delivered an interpretable model with at most ten features and a short list of properties that emerged from several independent configurations and hold on independent sequences.
The lists contain known biology, recovered without human prompting: e.g., from simple positional features for pausing, and from LLM-driven feature generators that were not told the factor's name for eIF4E.
Beyond the known biology, every task also yielded at least one lead of its own: a threshold to construct for nucleation, a translocation mechanism shared with \textit{E.~coli} for pausing, a coupling of codon optimality to eIF4E recruitment for eIF4E, and two candidate co-regulators binding the downstream intron for PTBP1.
Those are the starting points for the next experiment, computational or at the bench, and they were produced from sequence and labels alone.


\section*{Discussion}

SeqMaestro provides a framework for turning nucleotide sequence prediction problems into a search over explicit, human-readable biological hypotheses. Rather than committing in advance to a single representation of sequence or explaining a complex predictive model after it has been fitted, SeqMaestro searches across multiple interpretable representations, feature-generation strategies, and transparent model classes. Across the four biological questions considered here, this search produced compact predictive models and a small set of sequence properties that could be inspected directly and evaluated for robustness. The central contribution is therefore not simply an additional sequence classifier but an interface through which the broader ecosystem of interpretable machine learning can be applied to nucleotide sequence analysis. In this formulation, both the sequence property being measured and the predictive relationship in which it participates can form part of the resulting hypothesis.

The results also illustrate why no single interpretable representation is likely to be sufficient across sequence analysis problems. Positional representations are well suited to signals tied to specific locations relative to a landmark, whereas pattern-based representations provide a compact vocabulary for recurring local sequence elements. Programmatic features extend this vocabulary to quantities that combine regions, positions, counts, composition, or other sequence properties within a single interpretable computation. Importantly, the more expressive representation was not uniformly preferable. In some tasks, simple positional features were already sufficient to capture most of the predictive signal, whereas in others, programmatic features retained substantially more predictive information when models were reduced to a small number of features. This suggests that the choice of representation should itself be treated as part of the hypothesis search rather than as a preprocessing decision made before analysis.

The same argument applies to the transparent model class. Different model families expose different aspects of the relationship between a sequence property and the target. Sparse linear models provide additive effects with an explicit direction and magnitude; additive models can reveal gradients, thresholds, saturation, or reversals across the range of a feature; and decision trees can express conditional rules in which a sequence property matters only when another condition is satisfied. SeqMaestro therefore goes beyond identifying which features are important. A biological hypothesis may instead concern the shape of an effect, a threshold, or an interaction between interpretable sequence properties. This distinction is important because a ranked feature list can identify a predictive signal while leaving much of its predictive logic unspecified. By keeping the fitted model itself interpretable, SeqMaestro allows that logic to remain part of the scientific output.

Large language models play a deliberately restricted role in this process. They are used to propose programmatic features and, in some configurations, to refine those proposals using feedback from fitted models. They do not determine whether a feature is retained, whether it predicts the outcome, or whether it passes the reliability analysis. Those decisions are made using the observed data and fixed statistical procedures. The framework also records which information was exposed to the feature generator, enabling the withholding of identifying biological context when the goal is to distinguish discovery from the reproduction of prior knowledge.

SeqMaestro is intended to generate predictive and testable hypotheses, not to establish causal mechanisms directly. A sequence property that is reproducibly associated with a label may itself be causal, may serve as a proxy for another sequence property, or may reflect a confounding aspect of how the dataset was constructed. None of the reliability tiers can distinguish these possibilities in general. The same limitation applies to any supervised analysis of observational sequence data and places particular importance on the construction of the prediction task. Several additional limitations define the current scope of the framework. The implementation considered here addresses binary classification, although regression, multiclass, and multilabel settings are natural extensions. Stability under resampling of one dataset is weaker evidence than replication across independent datasets, conditions, species, or laboratories. Finally, the experiments in this study are intended to evaluate interpretable hypothesis search rather than to benchmark SeqMaestro against the full range of modern sequence foundation models or highly optimized deep sequence predictors. The gradient-boosted models provide a reference within the feature spaces considered here, but they do not constitute such a comparison.

These limitations also suggest several directions for extension. Additional transparent model families could broaden the forms of predictive relationships that can be expressed. Reliability analysis could incorporate replication across independent datasets as a further level of evidence. More broadly, the same architecture could support iterative experimental workflows in which hypotheses generated from one dataset are tested,, and the resulting measurements serve as the starting point for the next search. SeqMaestro thus provides a route from sequence prediction toward a more explicit form of computational hypothesis generation, in which the output of machine learning is not only a prediction or an attribution score, but a compact, computable statement about nucleotide sequence that can be inspected, challenged, and tested.

\bibliographystyle{unsrtnat}
\bibliography{references}

@article{rudin2019interpretable,
  author  = {Rudin, Cynthia},
  title   = {Stop explaining black box machine learning models for high stakes decisions and use
             interpretable models instead},
  journal = {Nature Machine Intelligence},
  volume  = {1},
  number  = {5},
  pages   = {206--215},
  year    = {2019},
  doi     = {10.1038/s42256-019-0048-x}
}

@inproceedings{lou2012intelligible,
  author    = {Lou, Yin and Caruana, Rich and Gehrke, Johannes},
  title     = {Intelligible models for classification and regression},
  booktitle = {Proceedings of the 18th ACM SIGKDD International Conference on Knowledge Discovery
               and Data Mining},
  pages     = {150--158},
  year      = {2012},
  publisher = {ACM},
  doi       = {10.1145/2339530.2339556},
}

@inproceedings{lou2013accurate,
  author    = {Lou, Yin and Caruana, Rich and Gehrke, Johannes and Hooker, Giles},
  title     = {Accurate intelligible models with pairwise interactions},
  booktitle = {Proceedings of the 19th ACM SIGKDD International Conference on Knowledge Discovery
               and Data Mining},
  pages     = {623--631},
  year      = {2013},
  publisher = {ACM},
  doi       = {10.1145/2487575.2487579},
}

@article{nori2019interpretml,
  author  = {Nori, Harsha and Jenkins, Samuel and Koch, Paul and Caruana, Rich},
  title   = {{InterpretML}: {A} unified framework for machine learning interpretability},
  journal = {arXiv},
  volume  = {1909.09223},
  year    = {2019},
  doi     = {10.48550/arXiv.1909.09223}
}

@inproceedings{chen2016xgboost,
  author    = {Chen, Tianqi and Guestrin, Carlos},
  title     = {{XGBoost}: {A} scalable tree boosting system},
  booktitle = {Proceedings of the 22nd ACM SIGKDD International Conference on Knowledge Discovery
               and Data Mining},
  pages     = {785--794},
  year      = {2016},
  publisher = {ACM},
  doi       = {10.1145/2939672.2939785},
}

@article{pedregosa2011scikit,
  author  = {Pedregosa, Fabian and Varoquaux, Ga{\"e}l and Gramfort, Alexandre and Michel, Vincent
             and Thirion, Bertrand and Grisel, Olivier and Blondel, Mathieu and Prettenhofer, Peter
             and Weiss, Ron and Dubourg, Vincent and Vanderplas, Jake and Passos, Alexandre and
             Cournapeau, David and Brucher, Matthieu and Perrot, Matthieu and Duchesnay, {\'E}douard},
  title   = {Scikit-learn: {M}achine learning in {Python}},
  journal = {Journal of Machine Learning Research},
  volume  = {12},
  pages   = {2825--2830},
  year    = {2011},
}

@article{huynh2026deft,
  author  = {Huynh, Nicolas and Kacprzyk, Krzysztof and Sheridan, Ryan and Bentley, David and
             van der Schaar, Mihaela},
  title   = {Interpretable {DNA} sequence classification via dynamic feature generation in decision
             trees},
  journal = {arXiv},
  volume  = {2604.12060},
  year    = {2026},
  note    = {Proceedings of the 29th International Conference on Artificial Intelligence and
             Statistics (AISTATS)},
  doi     = {10.48550/arXiv.2604.12060},
}

@article{nojima2015netseq,
  author  = {Nojima, Takayuki and Gomes, Tom{\'a}s and Grosso, Ana Rita Fialho and Kimura, Hiroshi
             and Dye, Michael J. and Dhir, Somdutta and Carmo-Fonseca, Maria and Proudfoot,
             Nicholas J.},
  title   = {Mammalian {NET}-seq reveals genome-wide nascent transcription coupled to {RNA}
             processing},
  journal = {Cell},
  volume  = {161},
  number  = {3},
  pages   = {526--540},
  year    = {2015},
  doi     = {10.1016/j.cell.2015.03.027},
}

@article{veronezi2024nucleation,
  author  = {Veronezi, Giovana M. B. and Ramachandran, Srinivas},
  title   = {Nucleation and spreading maintain {Polycomb} domains every cell cycle},
  journal = {Cell Reports},
  volume  = {43},
  number  = {4},
  pages   = {114090},
  year    = {2024},
  doi     = {10.1016/j.celrep.2024.114090}
}

@article{mendenhall2010gcrich,
  author  = {Mendenhall, Eric M. and Koche, Richard P. and Truong, Thanh and Zhou, Vicky W. and
             Issac, Biju and Chi, Andrew S. and Ku, Manching and Bernstein, Bradley E.},
  title   = {{GC}-rich sequence elements recruit {PRC2} in mammalian {ES} cells},
  journal = {PLoS Genetics},
  volume  = {6},
  number  = {12},
  pages   = {e1001244},
  year    = {2010},
  doi     = {10.1371/journal.pgen.1001244},
}

@article{shah2013variable,
  author  = {Shah, Rajen D. and Samworth, Richard J.},
  title   = {Variable selection with error control: {A}nother look at stability selection},
  journal = {Journal of the Royal Statistical Society Series B: Statistical Methodology},
  volume  = {75},
  number  = {1},
  pages   = {55--80},
  year    = {2013},
  doi     = {10.1111/j.1467-9868.2011.01034.x}
}

@article{faletto2022cluster,
  author  = {Faletto, Gregory and Bien, Jacob},
  title   = {Cluster stability selection},
  journal = {arXiv},
  volume  = {2201.00494},
  year    = {2022},
  doi     = {10.48550/arXiv.2201.00494}
}

@article{mann1947test,
  author  = {Mann, H. B. and Whitney, D. R.},
  title   = {On a test of whether one of two random variables is stochastically larger than the
             other},
  journal = {The Annals of Mathematical Statistics},
  volume  = {18},
  number  = {1},
  pages   = {50--60},
  year    = {1947},
  doi     = {10.1214/aoms/1177730491}
}

@article{phipson2010permutation,
  author  = {Phipson, Belinda and Smyth, Gordon K.},
  title   = {Permutation {P}-values should never be zero: {C}alculating exact {P}-values when permutations are randomly drawn},
  journal = {Statistical Applications in Genetics and Molecular Biology},
  volume  = {9},
  number  = {1},
  pages   = {Article 39},
  year    = {2010},
  doi     = {10.2202/1544-6115.1585}
}

@article{benjamini1995controlling,
  author  = {Benjamini, Yoav and Hochberg, Yosef},
  title   = {Controlling the false discovery rate: {A} practical and powerful approach to multiple testing},
  journal = {Journal of the Royal Statistical Society Series B: Statistical Methodology},
  volume  = {57},
  number  = {1},
  pages   = {289--300},
  year    = {1995},
  doi     = {10.1111/j.2517-6161.1995.tb02031.x}
}

@article{benjamini2001control,
  author  = {Benjamini, Yoav and Yekutieli, Daniel},
  title   = {The control of the false discovery rate in multiple testing under dependency},
  journal = {The Annals of Statistics},
  volume  = {29},
  number  = {4},
  pages   = {1165--1188},
  year    = {2001},
  doi     = {10.1214/aos/1013699998}
}

@article{storey2003statistical,
  author  = {Storey, John D. and Tibshirani, Robert},
  title   = {Statistical significance for genomewide studies},
  journal = {Proceedings of the National Academy of Sciences},
  volume  = {100},
  number  = {16},
  pages   = {9440--9445},
  year    = {2003},
  doi     = {10.1073/pnas.1530509100}
}

@book{westfall1993resampling,
  author    = {Westfall, Peter H. and Young, S. Stanley},
  title     = {Resampling-based multiple testing: {E}xamples and methods for {P}-value adjustment},
  publisher = {Wiley},
  address   = {New York},
  year      = {1993},
}

@article{holm1979simple,
 author = {Sture Holm},
 journal = {Scandinavian Journal of Statistics},
 number = {2},
 pages = {65--70},
 title = {A Simple Sequentially Rejective Multiple Test Procedure},
 volume = {6},
 year = {1979}
}

@article{lei2018distribution,
  author  = {Lei, Jing and G'Sell, Max and Rinaldo, Alessandro and Tibshirani, Ryan J. and
             Wasserman, Larry},
  title   = {Distribution-free predictive inference for regression},
  journal = {Journal of the American Statistical Association},
  volume  = {113},
  number  = {523},
  pages   = {1094--1111},
  year    = {2018},
  doi     = {10.1080/01621459.2017.1307116}
}

@article{hanley1982meaning,
  author  = {Hanley, James A. and McNeil, Barbara J.},
  title   = {The meaning and use of the area under a receiver operating characteristic ({ROC})
             curve},
  journal = {Radiology},
  volume  = {143},
  number  = {1},
  pages   = {29--36},
  year    = {1982},
  doi     = {10.1148/radiology.143.1.7063747}
}

@article{bourgon2010independent,
  author  = {Bourgon, Richard and Gentleman, Robert and Huber, Wolfgang},
  title   = {Independent filtering increases detection power for high-throughput experiments},
  journal = {Proceedings of the National Academy of Sciences},
  volume  = {107},
  number  = {21},
  pages   = {9546--9551},
  year    = {2010},
  doi     = {10.1073/pnas.0914005107},
}

@article{farcas2012kdm2b,
  author  = {Farcas, Anca M and Blackledge, Neil P and Sudbery, Ian and Long, Hannah K and McGouran, Joanna F and Rose, Nathan R and Lee, Sheena and Sims, David and Cerase, Andrea and Sheahan, Thomas W and Koseki, Haruhiko and Brockdorff, Neil and Ponting, Chris P and Kessler, Benedikt M and Klose, Robert J},
  title   = {{KDM2B} links the {Polycomb} Repressive Complex 1 ({PRC1}) to recognition of {CpG} islands},
  journal = {eLife},
  volume  = {1},
  year    = {2012},
  pages   = {e00205},
  doi     = {10.7554/elife.00205}
}

@article{blackledge2014variant,
  author  = {Blackledge, Neil P. and Farcas, Anca M. and Kondo, Takashi and King, Hamish W. and McGouran, Joanna F. and Hanssen, Lars L.P. and Ito, Shinsuke and Cooper, Sarah and Kondo, Kaori and Koseki, Yoko and Ishikura, Tomoyuki and Long, Hannah K. and Sheahan, Thomas W. and Brockdorff, Neil and Kessler, Benedikt M. and Koseki, Haruhiko and Klose, Robert J.},
  title   = {Variant {PRC1} Complex-Dependent {H2A} Ubiquitylation Drives {PRC2} Recruitment and {Polycomb} Domain Formation},
  journal = {Cell},
  volume  = {157},
  number  = {6},
  pages   = {1445-1459},
  year    = {2014},
  doi     = {10.1016/j.cell.2014.05.004}
}

@article{oksuz2018capturing,
  author  = {Oksuz, Ozgur and Narendra, Varun and Lee, Chul-Hwan and Descostes, Nicolas and LeRoy, Gary and Raviram, Ramya and Blumenberg, Lili and Karch, Kelly and Rocha, Pedro P. and Garcia, Benjamin A. and Skok, Jane A. and Reinberg, Danny},
  title   = {Capturing the Onset of {PRC2}-Mediated Repressive Domain Formation},
  journal = {Molecular Cell},
  volume  = {70},
  number  = {6},
  pages   = {1149-1162.e5},
  year    = {2018},
  doi     = {10.1016/j.molcel.2018.05.023}
}

@article{reveron2018accurate,
  author  = {Reverón-Gómez, Nazaret and González-Aguilera, Cristina and Stewart-Morgan, Kathleen R. and Petryk, Nataliya and Flury, Valentin and Graziano, Simona and Johansen, Jens Vilstrup and Jakobsen, Janus Schou and Alabert, Constance and Groth, Anja},
  title   = {Accurate Recycling of Parental Histones Reproduces the Histone Modification Landscape during {DNA} Replication},
  journal = {Molecular Cell},
  volume  = {72},
  number  = {2},
  pages   = {239-249.e5},
  year    = {2018},
  doi     = {10.1016/j.molcel.2018.08.010}
}

@article{larson2014pause,
  author  = {Larson, Matthew H. and Mooney, Rachel A. and Peters, Jason M. and Windgassen, Tricia and Nayak, Dhananjaya and Gross, Carol A. and Block, Steven M. and Greenleaf, William J. and Landick, Robert and Weissman, Jonathan S.},
  title   = {A pause sequence enriched at translation start sites drives transcription dynamics in vivo},
  journal = {Science},
  volume  = {344},
  number  = {6187},
  pages   = {1042-1047},
  year    = {2014},
  doi     = {10.1126/science.1251871}
}

@article{vvedenskaya2014interactions,
  author  = {Vvedenskaya, Irina O. and Vahedian-Movahed, Hanif and Bird, Jeremy G. and Knoblauch, Jared G. and Goldman, Seth R. and Zhang, Yu and Ebright, Richard H. and Nickels, Bryce E.},
  title   = {Interactions between {RNA} polymerase and the ``core recognition element'' counteract pausing},
  journal = {Science},
  volume  = {344},
  number  = {6189},
  pages   = {1285-1289},
  year    = {2014},
  doi     = {10.1126/science.1253458}
}

@article{gajos2021conserved,
  author  = {Gajos, Martyna and Jasnovidova, Olga and van Bömmel, Alena and Freier, Susanne and Vingron, Martin and Mayer, Andreas},
  title   = {Conserved {DNA} sequence features underlie pervasive {RNA} polymerase pausing},
  journal = {Nucleic Acids Research},
  volume  = {49},
  number  = {8},
  pages   = {4402-4420},
  year    = {2021},
  doi     = {10.1093/nar/gkab208}
}

@article{sheridan2019widespread,
  author  = {Sheridan, Ryan M. and Fong, Nova and D'Alessandro, Angelo and Bentley, David L.},
  title   = {Widespread Backtracking by {RNA} {Pol} {II} Is a Major Effector of Gene Activation, 5' Pause Release, Termination, and Transcription Elongation Rate},
  journal = {Molecular Cell},
  volume  = {73},
  number  = {1},
  pages   = {107-118.e4},
  year    = {2019},
  doi     = {10.1016/j.molcel.2018.10.031},
}

@article{llorian2010position,
  author  = {Llorian, Miriam and Schwartz, Schraga and Clark, Tyson A and Hollander, Dror and Tan, Lit-Yeen and Spellman, Rachel and Gordon, Adele and Schweitzer, Anthony C and de la Grange, Pierre and Ast, Gil and Smith, Christopher W J},
  title   = {Position-dependent alternative splicing activity revealed by global profiling of alternative splicing events regulated by {PTB}},
  journal = {Nature Structural \& Molecular Biology},
  volume  = {17},
  number  = {9},
  pages   = {1114-1123},
  year    = {2010},
  doi     = {10.1038/nsmb.1881}
}

@article{xue2009genome,
  author  = {Xue, Yuanchao and Zhou, Yu and Wu, Tongbin and Zhu, Tuo and Ji, Xiong and Kwon, Young-Soo and Zhang, Chao and Yeo, Gene and Black, Douglas L. and Sun, Hui and Fu, Xiang-Dong and Zhang, Yi},
  title   = {Genome-wide Analysis of {PTB}-{RNA} Interactions Reveals a Strategy Used by the General Splicing Repressor to Modulate Exon Inclusion or Skipping},
  journal = {Molecular Cell},
  volume  = {36},
  number  = {6},
  pages   = {996-1006},
  year    = {2009},
  doi     = {10.1016/j.molcel.2009.12.003},
}

@article{gueroussov2015alternative,
  author  = {Gueroussov, Serge and Gonatopoulos-Pournatzis, Thomas and Irimia, Manuel and Raj, Bushra and Lin, Zhen-Yuan and Gingras, Anne-Claude and Blencowe, Benjamin J.},
  title   = {An alternative splicing event amplifies evolutionary differences between vertebrates},
  journal = {Science},
  volume  = {349},
  number  = {6250},
  pages   = {868-873},
  year    = {2015},
  doi     = {10.1126/science.aaa8381}
}

@article{altschul1990basic,
  title={Basic local alignment search tool},
  author={Altschul, Stephen F and Gish, Warren and Miller, Webb and Myers, Eugene W and Lipman, David J},
  journal={Journal of Molecular Biology},
  volume={215},
  number={3},
  pages={403--410},
  year={1990}
}

@article{moeckel2024survey,
  title     = "A survey of k-mer methods and applications in bioinformatics",
  author    = "Moeckel, Camille and Mareboina, Manvita and Konnaris, Maxwell A
               and Chan, Candace S Y and Mouratidis, Ioannis and Montgomery,
               Austin and Chantzi, Nikol and Pavlopoulos, Georgios A and
               Georgakopoulos-Soares, Ilias",
  journal   = "Computational and Structural Biotechnology Journal",
  volume    =  23,
  pages     = "2289--2303",
  year      =  2024
}

@article{bailey2006meme,
  title={{MEME}: {D}iscovering and analyzing {DNA} and protein sequence motifs},
  author={Bailey, Timothy L and Williams, Nadya and Misleh, Chris and Li, Wilfred W},
  journal={Nucleic Acids Research},
  volume={34},
  number={suppl\_2},
  pages={W369--W373},
  year={2006}
}

@article{ghandi2014enhanced,
  title={Enhanced regulatory sequence prediction using gapped k-mer features},
  author={Ghandi, Mahmoud and Lee, Dongwon and Mohammad-Noori, Morteza and Beer, Michael A},
  journal={PLoS Computational Biology},
  volume={10},
  number={7},
  pages={e1003711},
  year={2014}
}

@article{zhou2015predicting,
  title={Predicting effects of noncoding variants with deep learning--based sequence model},
  author={Zhou, Jian and Troyanskaya, Olga G},
  journal={Nature Methods},
  volume={12},
  number={10},
  pages={931--934},
  year={2015}
}

@article{avsec2021effective,
  title={Effective gene expression prediction from sequence by integrating long-range interactions},
  author={Avsec, {\v{Z}}iga and Agarwal, Vikram and Visentin, Daniel and Ledsam, Joseph R and Grabska-Barwinska, Agnieszka and Taylor, Kyle R and Assael, Yannis and Jumper, John and Kohli, Pushmeet and Kelley, David R},
  journal={Nature Methods},
  volume={18},
  number={10},
  pages={1196--1203},
  year={2021},
  doi="https://doi.org/10.1038/s41592-021-01252-x"
}

@article{dalla2025nucleotide,
  title={Nucleotide {Transformer}: {B}uilding and evaluating robust foundation models for human genomics},
  author={Dalla-Torre, Hugo and Gonzalez, Liam and Mendoza-Revilla, Javier and Lopez Carranza, Nicolas and Grzywaczewski, Adam Henryk and Oteri, Francesco and Dallago, Christian and Trop, Evan and De Almeida, Bernardo P and Sirelkhatim, Hassan and others},
  journal={Nature Methods},
  volume={22},
  number={2},
  pages={287--297},
  year={2025},
  doi="https://doi.org/10.1038/s41592-024-02523-z"
}

@article{adebayo2018sanity,
  title={Sanity checks for saliency maps},
  author={Adebayo, Julius and Gilmer, Justin and Muelly, Michael and Goodfellow, Ian and Hardt, Moritz and Kim, Been},
  journal={Advances in Neural Information Processing Systems},
  volume={31},
  year={2018}
}

@inproceedings{ghorbani2019interpretation,
  author       = {Amirata Ghorbani and
                  Abubakar Abid and
                  James Y. Zou},
  title        = {Interpretation of Neural Networks Is Fragile},
  booktitle    = {Proceedings of the {AAAI} Conference on Artificial Intelligence},
  volume       = {33},
  pages        = {3681--3688},
  publisher    = {{AAAI} Press},
  year         = {2019},
  doi          = {10.1609/aaai.v33i01.33013681},
}

@article{chen2024applying,
  title={Applying interpretable machine learning in computational biology—pitfalls, recommendations and opportunities for new developments},
  author={Chen, Valerie and Yang, Muyu and Cui, Wenbo and Kim, Joon Sik and Talwalkar, Ameet and Ma, Jian},
  journal={Nature Methods},
  volume={21},
  number={8},
  pages={1454--1461},
  year={2024}
}

@article{burge1997prediction,
  title={Prediction of complete gene structures in human genomic {DNA}},
  author={Burge, Chris and Karlin, Samuel},
  journal={Journal of Molecular Biology},
  volume={268},
  number={1},
  pages={78--94},
  year={1997},
}

@ARTICLE{Roiuk2024,
  title     = "{eIF4E-independent} translation is largely {eIF3d-dependent}",
  author    = "Roiuk, Mykola and Neff, Marilena and Teleman, Aurelio A",
  journal   = "Nature Communications",
  volume    =  15,
  number    =  1,
  pages     = "6692",
  year      =  2024
}

@ARTICLE{Mignone2002,
  title     = "Untranslated regions of {mRNAs}",
  author    = "Mignone, Flavio and Gissi, Carmela and Liuni, Sabino and Pesole,
               Graziano",
  journal   = "Genome Biology",
  volume    =  3,
  number    =  3,
  pages     = "REVIEWS0004",
  year      =  2002,
  doi       = "10.1186/gb-2002-3-3-reviews0004"
}

@ARTICLE{Shabalina2004,
  title     = "Comparative analysis of orthologous eukaryotic {mRNAs}: {P}otential hidden functional signals",
  author    = "Shabalina, Svetlana A and Ogurtsov, Aleksey Y and Rogozin, Igor
               B and Koonin, Eugene V and Lipman, David J",
  journal   = "Nucleic Acids Research",
  volume    =  32,
  number    =  5,
  pages     = "1774--1782",
  year      =  2004
}

@ARTICLE{Pesole2001,
  title     = "Structural and functional features of eukaryotic {mRNA}
               untranslated regions",
  author    = "Pesole, G and Mignone, F and Gissi, C and Grillo, G and
               Licciulli, F and Liuni, S",
  journal   = "Gene",
  volume    =  276,
  number    = "1-2",
  pages     = "73--81",
  year      =  2001
}

@ARTICLE{Acevedo2018,
  title     = "Changes in global translation elongation or initiation rates shape the proteome via the {Kozak} sequence",
  author    = "Acevedo, Julieta M and Hoermann, Bernhard and Schlimbach, Tilo
               and Teleman, Aurelio A",
  journal   = "Scientific Reports",
  volume    =  8,
  number    =  1,
  pages     = "4018",
  year      =  2018
}

@ARTICLE{Hia2019,
  title     = "Codon bias confers stability to human {mRNAs}",
  author    = "Hia, Fabian and Yang, Sheng Fan and Shichino, Yuichi and
               Yoshinaga, Masanori and Murakawa, Yasuhiro and Vandenbon, Alexis
               and Fukao, Akira and Fujiwara, Toshinobu and Landthaler, Markus
               and Natsume, Tohru and Adachi, Shungo and Iwasaki, Shintaro and
               Takeuchi, Osamu",
  journal   = "EMBO Reports",
  volume    =  20,
  number    =  11,
  pages     = "e48220",
  year      =  2019,
  doi       = "10.15252/embr.201948220"
}

@ARTICLE{Barrington2023,
  title     = "Synonymous codon usage regulates translation initiation",
  author    = "Barrington, Chloe L and Galindo, Gabriel and Koch, Amanda L and
               Horton, Emma R and Morrison, Evan J and Tisa, Samantha and
               Stasevich, Timothy J and Rissland, Olivia S",
  journal   = "Cell Reports",
  volume    =  42,
  number    =  12,
  pages     = "113413",
  year      =  2023,
  doi       = "10.1016/j.celrep.2023.113413"
}

@ARTICLE{Fong2022,
  title     = "The pausing zone and control of {RNA} polymerase {II} elongation by {Spt5}: {I}mplications for the pause-release model",
  author    = "Fong, Nova and Sheridan, Ryan M and Ramachandran, Srinivas and
               Bentley, David L",
  journal   = "Molecular Cell",
  volume    =  82,
  number    =  19,
  pages     = "3632--3645.e4",
  year      =  2022
}

@ARTICLE{Kang2018,
  title     = "{RNA} polymerase accommodates a pause {RNA} hairpin by global
               conformational rearrangements that prolong pausing",
  author    = "Kang, Jin Young and Mishanina, Tatiana V and Bellecourt, Michael
               J and Mooney, Rachel Anne and Darst, Seth A and Landick, Robert",
  journal   = "Molecular Cell",
  volume    =  69,
  number    =  5,
  pages     = "802--815.e5",
  year      =  2018
}

@ARTICLE{Ule2019,
  title     = "Alternative splicing regulatory networks: {F}unctions, mechanisms,
               and evolution",
  author    = "Ule, Jernej and Blencowe, Benjamin J",
  journal   = "Molecular Cell",
  volume    =  76,
  number    =  2,
  pages     = "329--345",
  year      =  2019
}

@ARTICLE{Hu2018,
  title     = "{PTB/nPTB}: {M}aster regulators of neuronal fate in mammals",
  author    = "Hu, Jing and Qian, Hao and Xue, Yuanchao and Fu, Xiang-Dong",
  journal   = "Biophysics Reports",
  volume    =  4,
  number    =  4,
  pages     = "204--214",
  year      =  2018
}

@ARTICLE{Hamid2017,
  title     = "A mechanism underlying position-specific regulation of alternative splicing",
  author    = "Hamid, Fursham M and Makeyev, Eugene V",
  journal   = "Nucleic Acids Research",
  volume    =  45,
  number    =  21,
  pages     = "12455--12468",
  year      =  2017
}

@ARTICLE{Millevoi2009,
  title     = "A physical and functional link between splicing factors promotes {pre-mRNA} 3' end processing",
  author    = "Millevoi, Stefania and Decorsi{\`e}re, Adrien and Loulergue,
               Clarisse and Iacovoni, Jason and Bernat, Sandra and Antoniou,
               Michael and Vagner, St{\'e}phan",
  journal   = "Nucleic Acids Research",
  volume    =  37,
  number    =  14,
  pages     = "4672--4683",
  year      =  2009,
}

\clearpage
\section*{Methods}

\subsection*{Overview}

SeqMaestro is given a set of nucleotide sequences, each labelled with one of two classes, together
with a short description of what those classes mean and how they were measured. It fits a range of
interpretable models to the sequences, records which properties of a sequence each model relies on,
and then measures how well each of those properties is supported. The result is a report: every model
that was fitted and how it scored, the small interpretable models presented feature by feature, and a panel
of reliability analysis of features and concepts found.

Since somebody who wants to score new sequences needs the most accurate small model, yet somebody designing an
experiment or follow-up investigation is interested in the properties that several independent methods converged
on reliably, SeqMaestro's report covers both of these parallel goals.

The work divides into three stages, shown in Figure~\ref{fig:method}. Figure~\ref{fig:flow}
in Appendix~\ref{app:features} builds on it by following individual features through the same
stages, whether they were derived from the data alone or proposed by the language model, up to
the point where they reach the report.

\begin{figure}[p]
\centering
\includegraphics[width=0.915\linewidth]{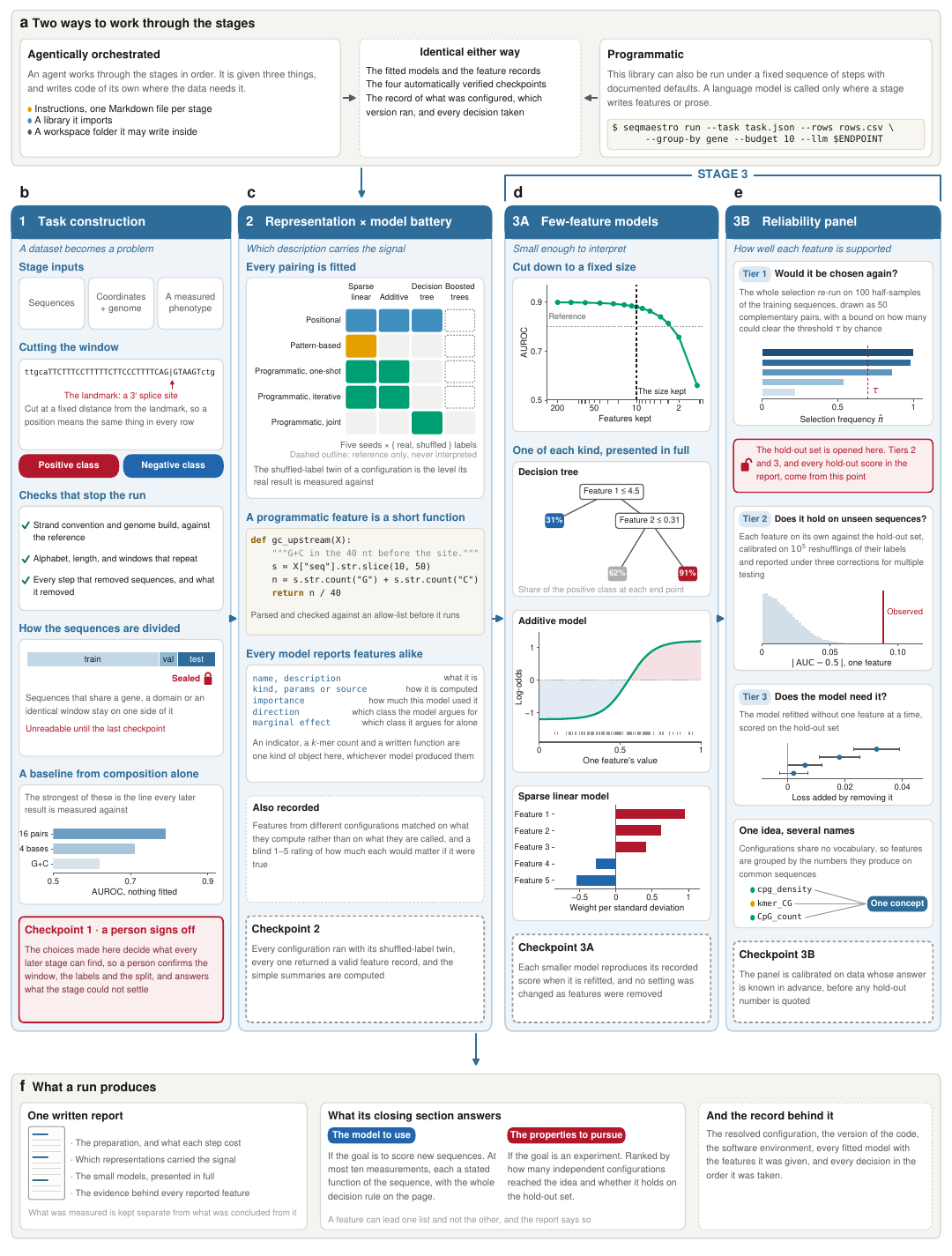}
\caption{\textbf{A SeqMaestro run.}
\textbf{(a)} The three stages are driven either by a tool-using AI agent, given written instructions for each stage, the SeqMaestro library, and a workspace folder, or run as a single command with sensible, documented defaults. Both routes follow these stages.
\textbf{(b)} Stage 1 turns delivered data into a prediction problem, based on user interaction: the choice of sequence windows, and where they are aligned; provenance and composition of the data is checked; every filtered or discarded sequence is accounted for; and sequence stratification and grouping are defined.
A simple composition-only model is recorded as a baseline. Finally, the user is asked to check and sign off this work at the checkpoint.
\textbf{(c)} Stage 2 fits every pairing of a sequence feature representation with a compatible model, including a shuffled-label control. Key properties of discovered features are recorded in a standardized format.
\textbf{(d)} Stage 3A reduces each configuration to a fixed number of features and records the best performing interpretable model of each kind.
\textbf{(e)} Stage 3B evaluates every candidate feature whether the selection would return it on another half of the data, whether it separates the classes on unseen sequences, and whether the model loses accuracy when the feature is individually discarded. The hold-out set is opened the first of those three questions is settled. The other two (and every hold-out score in the report) are evaluated at the end.
\textbf{(f)} A run produces a written report, which discusses both the best interpretable models, and the robust concepts/hypotheses discovered.}
\label{fig:method}
\end{figure}

\begin{enumerate}[leftmargin=1.6em,itemsep=0.3em,topsep=0.4em]
\item \textbf{Stage 1} turns the user-provided data into a prediction problem. It settles what is being predicted, from which part(s) of sequence, and how the sequences are divided between the training, validation, and hold-out sets.
\item \textbf{Stage 2} pairs each of the five sequence feature representations with every compatible model family, and fits all the combinations, forming a representation-model ``battery'' of configurations. This is executed and systematically analysed to gather evidence and identify the signal carried by the sequences.
\item \textbf{Stage 3} narrows the results to a tractable report. Stage 3A reduces each fitted model down to a small number of features (10, by default) and presents the best-performing small, interpretable models. Stage 3B, the reliability panel, then investigates whether every candidate feature would be found again if the analysis was repeated (tier 1), whether it separates the classes on sequences that weren't used to fit the models (tier 2), and, finally, whether the small model independently needs the feature.
\end{enumerate}

Three principles apply at every stage.

\paragraph{Sequences are held out until the end.} A test set portion of the sequences is set aside at the start and is unavailable to everything that follows. When an agent drives the run, the withholding is enforced by what is made available to it (e.g. the folder it reads after Stage 1 contains no test set or scores computed on it). The set is unsealed once, after the models of Stage 3A and the stability tier of Stage 3B have been fixed. A single final pass then computes every hold-out quantity in the report: the scores of the models,
and Tiers 2 and 3 of the panel that require measuring how a feature fares on unseen sequences.

\paragraph{A shuffled-label control is applied throughout.} Alongside each real fit we run a matched shuffled-label ``twin'', which differs in that the training labels have been
permuted. This control sets the minimum bar for any real data-driven result. It matters also for the programmatic features written by language models, as such a model may propose a plausible set of features regardless of the labels. Thus, it is important to evaluate how much better the model performs once the actual labels are used.

\paragraph{Each stage ends at a checkpoint.} A checkpoint is a list of conditions that are verified automatically, together with a short written summary for the user. The first checkpoint asks for the user's approval, as the choices made in Stage 1 affect what every later stage can possibly find, and hence corrections and direction by a domain expert user are especially important at this stage. The remaining three stages run without interruption, unless interrupted by the user.

Table~\ref{tab:system}, in Appendix~\ref{app:glance}, contains further details that supplement this section.

\subsection*{Stage 1: building the prediction task}

Three parts of SeqMaestro have to explain the data to a language model. They are the component that proposes features, the decision tree that proposes features while it grows, and the rating step noted below. All three are handed a short description of the dataset, which describes what is being predicted, what the sequences are, and optional additional points of note supplied by the user. The user may also choose to not reveal the nature of the label (e.g. to withhold the protein name) from the LLM feature generating components of SeqMaestro. Stage 1 ensures that the description is settled with the user's input.

Each sequence in the analysis requires a window of fixed length, set at a fixed distance from a landmark chosen for biological reasons (this might be a splice site, a start codon, the position at which some event was measured, or simply a midpoint of an interval when the sequence has no orientation).
Anchoring the window this way makes a given position correspond to the same kind of location in every sequence, as otherwise ``position 43'', for instance, would be ambiguous.
Where the window cannot be derived from every user-provided sequence, SeqMaestro counts what was lost and reports it separately for each class.
A window of fixed size applied to sequences of unequal length may affect the class balance -- length is seldom unrelated to the label, so a step that removes a third of one class and a twentieth of the other affects the task setup -- and it is crucial that this information is tracked and presented to the user.

SeqMaestro then checks the sequences themselves, before anything at all is fitted. A failing check here halts the run, and reverts to the user for extra data or clarification.
The first check looks a sample of windows up in the reference genome (where available), at the coordinates of the sequences, and verifies the strand convention, so the genome build and the strand orientation are established.
The second check covers the alphabet (A, T, G, C, U, and so on), the length, and whether any window appears to have erroneous duplicate copies.
Another check concerns what was filtered: every step that removes sequences has to record what it removed and how much of each class was lost.

Sequences are then assigned to the training, validation and hold-out (test) sets in groups.
A group is whatever makes two sequences dependent on one another, which might be a gene, a domain, a clonal family, etc., depending on the research problem.
This ensures fair comparison across the two classes.
The variable used for grouping is recorded and this grouping is programmatically enforced in the data splits throughout the SeqMaestro run.

Stage 1 also sets out a simple compositional baseline.
On the sequences and splits the models will use, we compute the fraction of each base, the fraction of G or C, the fraction of C or T, and the counts of sixteen adjacent base pairs, fitting each family with an unpenalised logistic regression, and pick the strongest of these.

The summary the user approves at the first checkpoint states all of the key decision points discussed above with a note on what would change depending on the user's answer.
In agentic mode, the user may repeatedly provide additional data, or make clarifications, before the stage is re-run and a final study design settled.

\subsection*{Stage 2: fitting the representation-model battery}

We call one feature source paired with one model family a configuration.
The three feature families introduced earlier appear in five variants, and pairing those with the model families that can consume them gives all the possible configurations listed in Table~\ref{tab:system}a.
Positional and pattern-based features are defined (enumerated) in advance from the training sequences.
Programmatic features are written by a language model, in the three ways our framework supports.
It can propose them all at once, refine them over several rounds, or generate them while a model is being fitted.

For the first two of those, the model sees the description of the task and a sample of labelled training sequences balanced between the classes.
It returns candidates as Python functions, each with a name, a sentence describing what it computes and a sentence arguing why it might matter.
When features are refined over rounds, each round shows the model what the fitted model made of the features proposed so far, such that improved proposals can be made.
When they are generated during fitting, the tree asks for candidates at each node as it grows, so a feature can be tailored to the sequences that reach that point in the tree.
All generated code is parsed and checked against an allow-list before it runs, as Appendix~\ref{app:details} describes.

The battery uses four model families. Three of them are intrinsically interpretable, and the fourth (XGBoost) is present as a point of black-box comparison.

\begin{itemize}[leftmargin=1.2em,itemsep=0.25em,topsep=0.35em]
\item \textbf{Sparse linear.} $L_1$-penalised logistic regression on standardised columns \citep{pedregosa2011scikit}. The magnitude of a coefficient says how much the model leans on a feature, while its sign says which class it argues for.
\item \textbf{Additive.} An explainable boosting machine \citep{lou2012intelligible,lou2013accurate,nori2019interpretml}, which fits one curve per feature by boosting over the columns in rotation. The shape of a curve shows how the effect changes across the range of values the feature takes.
\item \textbf{Decision tree.} A binary tree with a limit on depth, a cap on the number of leaves, and a requirement that every leaf hold at least one per cent of the training rows. Where features are generated during fitting, this is the adaptive tree of \citet{huynh2026deft}, which asks a language model for candidates at each node.
\item \textbf{Gradient-boosted trees} \citep{chen2016xgboost}, included to show what an accurate black-box model achieves on the same dataset.
\end{itemize}

All four are fitted with balanced class weights, which reduces to the ratio \texttt{scikit-learn} defines (an already balanced dataset is unaffected).
Each configuration is fitted at a configurable number (default five) random seeds under the real labels, and again with permuted labels.

Every configuration's features, as used by its model, are recorded in a common format to facilitate the subsequent analysis stages. 
Each contains the details of what it computes, how heavily the fitted model leaned on it (feature importance), which class the model uses it to argue for (direction), and how far it separates the classes on its own (marginal effect).

Every reported feature is also shown to a language model on its own and rated for how much it would matter if it were true, with features from the permuted fits mixed into the same panel to give that rating a baseline to be measured against, providing a supplementary LLM rating of each feature.

Appendix~\ref{app:details} gives the fields of the feature record and the matching procedure in full detail.

\subsection*{Stage 3A: reducing to a small model}

Each configuration is then cut down to a model of at most $N$ features, with $N$ chosen in advance and set to ten in all the work reported here.
The reduction repeatedly drops whichever features the fitted model leans on least and refits on what remains.
It begins from the features that model actually used, and applies the same feature reduction algorithm everywhere.
We record how validation performance changes at every step, so a reader can see the effect of reducing the feature count.
Appendix~\ref{app:details} elaborates on the details of this procedure.

For one configuration from each of the three interpretable families, the reduced model is then stored to be presented to the user in full.
The tree appears with its features, its thresholds, the number of sequences reaching each node and the proportion of the positive class there.
The additive model appears as one curve per feature above a histogram showing where the sequences actually fall.
The sparse linear model appears as signed weights per standard deviation, with a mark on any feature the model uses in the opposite direction compared to the direction the feature points to in isolation.

As part of a checkpoint, each presented model is refitted from the run's records, and has to reproduce the score already recorded for it, to ensure nothing was mangled during this stage.

\subsection*{Stage 3B: the reliability panel}

While Stage 3A focuses on best-performing yet small, interpretable models, Stage 3B goes beyond measuring performance, but rather aims to establish the robustness of the proposed concepts.
That is, for each concept, is this a property of the biology that another study of the same kind would find again?
The reliability panel supplies analysis and does not fit any additional models.
It leaves the models of Stage 3A untouched, and gives its results as curves across a range of thresholds, so a reader can see the whole trade-off across cut-offs.
It covers every feature used by the interpretable model configurations.
Table~\ref{tab:system}b summarises this, and Appendix~\ref{app:reliability} covers the methodology in more detail.
The first of its three questions (Tier 1) is answered on the training and validation sequences alone.
The hold-out set is opened for the second and third (Tiers 2 and 3), and that one ``opening'' also supplies every hold-out score in the report.

Features that barely vary are removed first, and those that remain are grouped by how strongly they correlate with one another.
This clustering is valuable because a feature and a near-duplicate of it can ``divide the attention'' of any selection procedure between them, so each of the three kinds of evidence below is reported for the individual feature and again for the cluster it belongs to \citep{faletto2022cluster}.

\paragraph{Tier 1: Would this feature turn up again?} We re-run the procedure that produced the reported model on 100 half-samples of the training and validation sequences, drawn as 50 complementary pairs, and record the proportion of those half-samples in which each feature appears \citep{shah2013variable}.
For any threshold above one half, the expected number of features clearing it whose true selection probability is no better than chance is bounded, and the bound depends on how many features were available and how many a typical fit selects.

\paragraph{Tier 2: Does the feature separate the classes where nothing was fitted?} For each candidate we compute how far its area under the curve departs from one half, using that feature alone as a score.
This is the rank-sum statistic of \citet{mann1947test}, which requires no assumption about the shape of the feature and applies equally to indicators, counts and continuous quantities.
It is computed on the hold-out set and calibrated against $10^5$ reshufflings of their labels \citep{phipson2010permutation}, then reported under three corrections for multiple testing \citep{benjamini1995controlling,benjamini2001control,westfall1993resampling}, alongside a direct estimate of how many of the calls made at each threshold would have been made with no signal present at all \citep{storey2003statistical}.

\paragraph{Tier 3: Does the model need the feature?} We refit the model without one of its features at a time, using the procedure and the sequences that produced it, and compare the two by the change in log loss on the hold-out set \citep{lei2018distribution}.
Where a correlation cluster contributes more than one feature to a model, the whole cluster is also removed at once.
A feature that costs little when removed suggests that it is \emph{substitutable} (rather than uninformative), and the cluster it belongs to reveals what may be standing in for it.

The panel also runs two sense-checks, on setups whose answer is known in advance, to verify that its results can be trusted.
Firstly, if one replaces the labels with a random permutation, the panel should claim nothing at any threshold.
And secondly, if one simulates labels from five known feature columns and those five should stand at the top of the selection frequencies and be found on the hold-out sequences.

\subsection*{The report}

The report is assembled from stored results as the final step of a SeqMaestro run.
Numbers arrive through tables generated from the run results.
Explanations that always apply, such as how to look at a curve from an additive model or what a $q$-value means, are included unchanged in every report.

The report is also designed to keep the two following kinds of statement apart.
Anything outside a marked block is either a number or a description of what was done, and can be traced to a table or a figure.
Anything that is written by the AI agent as advice or interpretation (rather than description of fact) is placed inside a marked block.
The closing recommendations are one such block, for example.

\subsection*{Two ways to run SeqMaestro}

An AI agent with access to a shell and a filesystem can work through the SeqMaestro stages in order.
It is given three things:
\begin{itemize}
      \item The SeqMaestro library, which is everything described above, packaged.
      \item A set of written instructions, one file per stage, loaded when that stage
      begins.
      \item The third is the working folder that it may write inside.
\end{itemize}
It is not given the hold-out sequences until the point in Stage 3B at which they are required, after every model and the stability tier have been fixed.
Within those limits it writes code of its own, where needed, typically in Stage 1, as that is where one has to make sense of an unfamiliar file format, or make dynamic decisions like noticing that a harmless-looking filter removes a large number of sequences of one class.
The design therefore leaves open how the agent gets there, as long as the final user-mediated checkpoint is executed.
Appendix~\ref{app:agent} describes the instructions and the checkpoints, with an example.

SeqMaestro can also run under a fixed sequence of steps with documented defaults.
The user supplies the sequences, the description of the task, the variable to group on, and connection details for a language model, which, here, is called only at the programmatic feature configurations, at a number of predefined checkpoints, and for the final report-writing stage.
This route produces the same fitted models, records and checkpoints.
However, it cannot ``improvise'', as it is a naturally less flexible approach.

\subsection*{Implementation}

SeqMaestro is written in Python.
Sparse logistic regression, decision trees, the code that makes the splits and the code that computes the scores come from scikit-learn \citep{pedregosa2011scikit}, the
additive model from InterpretML \citep{nori2019interpretml}, the comparison model from XGBoost \citep{chen2016xgboost}, and the adaptive feature-generating tree from the implementation released with \citet{huynh2026deft}.

The results reported here came from the agent-driven route.
The agent had shell and filesystem access and was driven by GPT-5.6 Sol.
The two feature proposers, the proposer the adaptive tree calls at each node, and the rating step all used GPT-5.4 through Azure OpenAI.

Appendix~\ref{app:compute} gives the computational requirements of a run, in wall-clock time and in tokens.

Scores are reported with percentile bootstrap intervals from 200 resamples of the evaluation set, the same number for every dataset so that intervals can be compared between them. Any threshold needed for an accuracy-style measure is chosen on the validation set and applied unchanged to the hold-out set.
Each run records the command it was started with, the arguments after defaults were resolved, the version of the code, and a full description of the software environment, among other provenance metadata.

\clearpage
\appendix
\renewcommand{\thesection}{\Alph{section}}
\setcounter{section}{0}
\makeatletter
\renewcommand{\@seccntformat}[1]{\ifnum\pdfstrcmp{#1}{section}=0 Appendix~\fi\csname the#1\endcsname\quad}
\makeatother
\renewcommand{\contentsname}{Appendices}
\setcounter{tocdepth}{1}
\renewcommand{\cftsecpresnum}{Appendix~}
\setlength{\cftsecnumwidth}{6.2em}
\renewcommand{\cftsecfont}{\normalfont}
\renewcommand{\cftsecpagefont}{\normalfont}
\renewcommand{\cftsecleader}{\cftdotfill{\cftdotsep}}
\tableofcontents
\bigskip

\section{The system at a glance}
\label{app:glance}

Three inventories, collected here so the Methods can point at them and carry on. Panel (a) lists the
configurations the battery fits. Panel (b) lists the kinds of evidence the reliability panel
produces. Panel (c) lists the written instructions an agent driving a run is given, beside what each
checkpoint verifies before the run continues.

\vspace{0.6em}
{\footnotesize
\setlength{\parindent}{0pt}
\setlength{\tabcolsep}{4pt}
\renewcommand{\arraystretch}{1.25}

\captionof{table}{\textbf{What a SeqMaestro run is made of.}
\textbf{(a)} The battery. Thirteen configurations, each a pairing of a feature source with a model
class, every one fitted at five seeds under real labels and again under shuffled training labels.
\textbf{(b)} The discovery reliability panel. Every quantity is reported for the individual feature and
again for the group of correlated features it belongs to, and every threshold is reported as a curve
and not as a single verdict. Appendix~\ref{app:reliability} gives the statistics in full.
\textbf{(c)} What an agent driving a run is given, and what each checkpoint verifies before the run
continues. The agent writes code of its own inside the folder for the run, so a checkpoint settles
what the run must contain rather than the route taken to it.}
\label{tab:system}

\vspace{0.7em}
\begin{minipage}{\linewidth}
{\raggedright \textbf{(a)~~The representation and model battery}\par}
\vspace{0.3em}
\begin{tabular}{@{}L{2.55cm}L{3.15cm}L{6.35cm}L{3.85cm}@{}}
\toprule
\textbf{Feature source} & \textbf{One feature is} & \textbf{How the candidates are obtained} &
\textbf{Model classes fitted} \\
\midrule
Positional &
An indicator for one base at one position &
Listed in advance, four per position of the window &
Sparse linear, additive, tree, boosted \\
Pattern-based &
The number of times a short subsequence occurs &
Listed in advance, every $k$-mer from $k=1$ to $6$ seen in the training sequences, alongside the
positional features &
Sparse linear, boosted \\
Programmatic, one-shot &
A short Python function of the sequence &
Proposed once by a language model shown the task description and a class-balanced sample of labelled
training rows &
Sparse linear, additive, boosted \\
Programmatic, iterative &
As above &
Proposed over five rounds, the first blind and each later one shown what the fitted model made of the
features proposed so far &
Sparse linear, additive, boosted \\
Programmatic, joint &
As above &
Proposed at each node while the tree is grown, so a feature can be conditioned on the rows that reach
that node &
Decision tree \\
\bottomrule
\end{tabular}

\vspace{0.45em}
{\footnotesize The additive model is not fitted on pattern-based features, because it boosts over
columns in rotation, and on a $k$-mer vocabulary that costs far more than the comparison is worth. Boosted trees are present as a reference for what an accurate model nobody can inspect
achieves on the same columns, and their features are recorded and never interpreted.\par}
\end{minipage}

\vspace{1.4em}
\begin{minipage}{\linewidth}
{\raggedright \textbf{(b)~~The discovery reliability panel}\par}
\vspace{0.3em}
\begin{tabular}{@{}L{2.35cm}L{3.55cm}L{6.15cm}L{3.85cm}@{}}
\toprule
\textbf{Step} & \textbf{The question it answers} & \textbf{How it is computed} &
\textbf{What clearing a threshold says} \\
\midrule
\textbf{Before}\newline filter and group &
Which candidates stand in for one another &
Absolute Spearman correlation between feature columns, complete linkage cut at $0.8$, on the training
and validation sequences. No label is consulted &
The features in a group carry the same information on these sequences \\
\textbf{Tier 1}\newline stability &
Would the selection return this feature on another half of the same data &
The share of 100 half-samples that select it, drawn as 50 complementary pairs, reported over a range
of thresholds beside a bound on how many features could clear each one by chance &
The model did not depend on which half of the data it happened to see \\
\textbf{Tier 2}\newline association &
Does the feature separate the classes where nothing was fitted &
How far $|\mathrm{AUC}-1/2|$ departs from zero for that feature used alone as a score, on the
hold-out set, against $10^5$
reshufflings of their labels. Reported under three multiplicity corrections and a direct estimate
of the false discovery proportion &
The association is present in sequences no part of the analysis was fitted on \\
\textbf{Tier 3}\newline necessity &
Does the reported model lose accuracy without this feature &
The paired difference in log loss against the same model refitted without it, on the hold-out
sequences, one-sided with a sign-flip check and corrected across the ten features &
Nothing else in the model substitutes for what it carries \\
\bottomrule
\end{tabular}
\end{minipage}

\vspace{1.4em}
\begin{minipage}{\linewidth}
{\raggedright \textbf{(c)~~The instructions an agent is given, and the checkpoint that closes each
stage}\par}
\vspace{0.3em}
\begin{tabular}{@{}L{2.85cm}L{0.75cm}L{5.45cm}L{6.35cm}@{}}
\toprule
\textbf{Instructions} & \textbf{Stage} & \textbf{What it must decide and record} &
\textbf{What its checkpoint will not pass} \\
\midrule
\texttt{task-} \texttt{construction.md} & 1 &
The window and its landmark, the label rule, every step that removed rows and what it removed from
each class, the grouping variable, and the questions only the data owner can settle &
Rows discarded with no record of what was removed, a group appearing on more than one side of the
split, or a description stating a fact listed as withheld. \textbf{A person signs this one off} \\
\texttt{battery.md} & 2 &
Which configurations were run, the seeds, and the wording every proposal was made under &
A configuration with no shuffled-label twin, a feature record that fails validation, or a
configuration whose stored feature values disagree with the features it reported \\
\texttt{few-feature-} \texttt{models.md} & 3A &
The budget, the order features are dropped in, and the rule for breaking ties &
A smaller model whose refitted score does not reproduce the one already recorded, or any setting
tuned during the reduction \\
\texttt{reliability-} \texttt{panel.md} & 3B &
The thresholds, the number of half-samples and of permutations, and every point at which the
re-run selection departs from the original &
A panel that calls something when the labels are shuffled, a panel that fails to recover a planted
signal, or a hold-out label read before the models of Stage 3A and the stability tier were fixed \\
\texttt{reporting.md} & final &
Which configurations lead the document, and the separation of measured fact from interpretation &
A number in the prose that appears in no included table, or an interpretation outside a marked
block \\
\midrule
\multicolumn{4}{@{}L{16.6cm}@{}}{\emph{Two further files are read at every stage rather than at
one.} \texttt{task-description.md} covers the single description every language model in the run
sees, and what is kept from it. \texttt{run-record.md} covers the log of decisions, the software
environment and the version of the code that together make a run repeatable.} \\
\bottomrule
\end{tabular}
\end{minipage}

}

\section{The battery and the reduction, in detail}
\label{app:details}

\subsection*{Executing code that a language model wrote}

The method works by having a language model write feature functions and then running them, and the
prompts contain rows of the user's data, so a maliciously constructed input file would otherwise be a
route from data to code execution. Generated code passes three checks before it runs. Its syntax tree
is inspected first: imports must come from a short list, attribute names beginning with a double
underscore are refused, references to \texttt{open}, \texttt{eval}, \texttt{exec}, \texttt{compile}
and \texttt{getattr} are refused, and nothing may appear at the top level of the file except imports
and function definitions. The code is then executed against a minimal set of built-in functions with
a restricted import mechanism. Finally, the syntax tree that was inspected is compiled directly,
and the text is never parsed a second time, so what runs is what was checked. A function
that fails any of these is discarded and counted, so a prompt that has started producing unusable
code shows up as a shrinking number of candidates instead of passing unnoticed.

Together these measures make accidental or casual misuse difficult. An adversary with a Python
interpreter and enough patience is stopped only by isolating the process, which is a separate piece
of engineering that we have not attempted here.

\subsection*{What is recorded about every feature}

Each configuration reports the features its fitted model actually used, and all of them report in the
same format. Every record carries a name, a description, an indication of what kind of feature it is,
either its recipe or its source code, the importance the fitted model assigns it, a second importance
computed by shuffling that column and seeing how far accuracy falls, the class the model uses it to
argue for, and how far it separates the classes on its own with no model involved. That last quantity
is $2 \times \mathrm{AUC} - 1$ over the training rows against the real labels, SeqMaestro computes it and none of the
models does, so it means one thing across the entire battery.

The two importances look similar and are not interchangeable. A coefficient, a boosting gain and an
impurity reduction are different quantities in different units, so importances reported by different
model families can be compared only in rank order. The shuffling-based importance requires nothing
except a fitted model, some rows and a metric, so it is computed the same way everywhere, including
for the models that run in a separate software environment.

Every configuration also writes out the values of its reported features on one common set of
sequences, the validation set. Storing the values, and not the means of
recomputing them, keeps every later comparison to a matter of loading two columns of numbers, and it
means that code written by a language model is never executed outside the run that produced it.

\subsection*{Matching features between configurations}

Different configurations have no vocabulary in common. One writes \texttt{pos\_50\_A} for an
indicator that position 50 holds an adenine, another writes \texttt{pos\_50\_is\_A} for it, and a language model asked to describe the middle of a window might call it
\texttt{central\_base\_is\_a}. Nothing in SeqMaestro compares such names. Two features are considered
the same when the absolute Spearman rank correlation between the values they produce on the common
sequences reaches 0.8, with an exact match between normalised definitions taking precedence where one
exists. Rank correlation is the right comparison because a count and a fraction of the same thing are
related by a monotone transformation and should match perfectly, and the absolute value is taken
because a feature and its negation carry identical information.

Features from two configurations are paired by the assignment that maximises total similarity, rather
than by repeatedly taking the closest remaining pair, so the result does not depend on the order in
which features happen to be listed. A concept is then a connected group in the network whose links
are these matches. Because the grouping is transitive, a chain of borderline matches can join two
properties that a reader would want to keep apart. The clearest symptom of that is a concept
containing two features from the same configuration, and the report says when it has happened.

\subsection*{Rating how much a feature would matter}

Every reported feature is shown to a language model on its own, with no indication of which
configuration produced it, and rated from one to five on a single question: how biologically
significant or interesting would this be, if it were true. The conditional does the important work,
because a model cannot check whether a description is true and would give a confident answer anyway.
Features from the permuted-label fits enter the same panel and are indistinguishable within it, and
are separated out only when the ratings are summarised, so the rating scheme carries its own
comparison against noise. What it measures is closer to how a feature is phrased than to what
produced it, and no result in this paper depends on it.

\subsection*{Reducing a configuration to a fixed number of features}

\begin{algorithm}[H]
\SetAlgoLined
\DontPrintSemicolon
\KwIn{$F$, the features the configuration reported; the training rows; the budget $N$}
fit the configuration's model on $F$\;
\While{$|F| > N$}{
  rank $F$ by the importance the fitted model reports, in absolute value\;
  \eIf{some features have zero importance}{
    drop all of them together\;
  }{
    \eIf{$|F| > 30$}{drop the weakest quarter\;}{drop the weakest one\;}
  }
  refit on what remains\;
}
\KwOut{the model fitted on the surviving $N$ features}
\caption{Reduction to a fixed number of features}
\label{alg:rfe}
\end{algorithm}

Algorithm~\ref{alg:rfe} leaves three things open, and we settle each of them once, for every
configuration and every dataset.

Removing a quarter of the surviving features at a time while more than thirty remain keeps the number
of refits proportional to the logarithm of the number of features it starts with rather than to that
number itself, and still leaves the last thirty steps, which are the ones anyone examines, at full
resolution.

Features with zero importance are removed together rather than a quarter at a time, because a zero
means the model ignored the feature entirely, and choosing an arbitrary quarter of the features a
model ignores would present an arbitrary choice as though it were a ranking.

Ties are broken by the most recent step at which a feature carried any importance, then by how much
it carried at that point, and finally by the order in which the configuration listed it, which makes
the result deterministic without any part of the rule consulting a label.

\section{The discovery reliability panel in full}
\label{app:reliability}

This appendix gives the statistics behind Stage 3B. The summary in the Methods and in
Table~\ref{tab:system}b is enough to follow the results. This appendix gives what somebody would need in
order to reimplement the panel, or to argue with it.

\subsection*{Notation}

One configuration is profiled at a time. Write $X \in \mathbb{R}^{n \times p}$ for its candidate
feature matrix and $y \in \{0,1\}^n$ for the labels, with $n_+$ positives and $n_-$ negatives. Write
$\hat S(\cdot)$ for the selection procedure that produced the configuration's reported model,
returning a set of $K$ feature names, with $K = 10$ throughout this work. Write $R \subset \{1,
\dots, p\}$ for the $K$ reported features. Tier 1 uses the training and validation rows. Tiers 2
and 3 require the use of the hold-out rows once, \emph{after} Stage 3A and Tier 1 are complete,
which is the point in a run where the hold-out sequences are used.

\subsection*{Filtering and grouping, before any label is consulted}

Candidates with zero variance, or with a non-zero prevalence below 1\% or above 99\%, are removed and
listed. That filter is applied again to the hold-out rows, so the candidate list is identical on
both sides. Neither the filter nor the grouping below consults a label, and that is why they can
precede the tests without spending any of the error budget those tests have to control
\citep{bourgon2010independent}.

The survivors are clustered on $1 - |\rho|$, where $\rho$ is the Spearman rank correlation between two
feature columns, by hierarchical clustering with complete linkage, cut so that every pair within a
cluster satisfies $|\rho| \ge 0.8$. Complete linkage makes that a guarantee about every pair
rather than only about the chain of links that joined them. The four indicators of one position are additionally
grouped as a position group, since they are mutually exclusive and correlation clustering will not
merge them. A cluster's representative is its member with the highest Tier 1 selection frequency.

We call these groups clusters for the rest of this appendix. They matter because near-duplicate
features split the vote of any selection procedure between them, so a feature can be reported by a fitted model and still appear unstable. Every tier is
therefore reported twice, once for the feature and once for the cluster containing it
\citep{faletto2022cluster}.

\subsection*{Tier 1: would the selection return this feature again}

The procedure is complementary pairs stability selection \citep{shah2013variable}. Let $D$ be the
training and validation rows, $n = |D|$. For $b = 1, \dots, B$ with $B = 50$, draw a random pair of
disjoint subsets $A_b, \bar{A}_b \subset D$, each of size $\lfloor n/2 \rfloor$, and run $\hat S$ on
each. The selection frequency of feature $j$ is
\[
  \hat{\pi}_j \;=\; \frac{1}{2B} \sum_{b=1}^{B}
  \Big[ \mathbf{1}\{ j \in \hat S(A_b) \} + \mathbf{1}\{ j \in \hat S(\bar{A}_b) \} \Big],
\]
so 100 half-samples in total. For a threshold $\tau \in (1/2, 1]$ the selected set is $S_\tau = \{ j :
\hat{\pi}_j \ge \tau \}$. Let $\hat q$ be the mean size of a selected set across the $2B$ fits, and
let $N$ be the set of features whose true probability of being selected on a half-sample is no greater
than $\hat q / p$, which is the level a feature would reach if the procedure were choosing at random.
Then
\[
  \mathbb{E}\, |S_\tau \cap N| \;\le\; \frac{\hat q^{\,2}}{(2\tau - 1)\, p}.
\]
The bound holds for any $B$ and needs no assumption about the selection procedure. Its usefulness
depends on how many candidates there are. With $p$ in the hundreds it is informative. With $p$ of
order ten it can exceed the number of candidates altogether and says nothing, and the report states
which of the two situations applies, so a reader is never left to work it out. The panel reports $|S_\tau|$ and the bound on the grid $\tau \in \{0.55, 0.60,
\dots, 1.00\}$, together with the smallest $\tau$ at which the bound falls below one and below one
half.

The same quantities are computed at the level of clusters, where a cluster counts as selected in a
half-sample if any member of it is, with $p$ the number of clusters and $\hat q$ the mean number of
clusters selected.

$\hat S$ mirrors the procedure that produced the reported model, with the hyper-parameters fixed to
those of that model, and must return approximately $K$ features on every call. For the sparse linear
model the penalty is chosen per call so that about $K$ coefficients are non-zero. For the additive
model and the tree, the top $K$ features by the importance the model reports are taken. Any point at which
this re-run departs from the original procedure is recorded with the results.

\subsection*{Tier 2: association on the hold-out set}

The statistic for feature $j$ is
\[
  T_j \;=\; \Big| \mathrm{AUC}_j - \tfrac{1}{2} \Big|, \qquad
  \mathrm{AUC}_j \;=\; \frac{U_j}{n_+ n_-},
\]
where $U_j$ is the rank-sum statistic of \citet{mann1947test} computed from that feature alone used
as a score, with average ranks for ties. Written this way the statistic is the area under the
receiver operating characteristic curve for that feature \citep{hanley1982meaning}. One statistic serves binary, count and continuous features, it
makes no assumption about the shape of a feature, and it detects monotone association only, which is
the main thing it cannot see.

The null is $B_2 = 10^5$ random permutations of the evaluation labels. Computing it naively would
require $p B_2$ rank operations. Instead each column is ranked once into $R \in \mathbb{R}^{n \times
p}$, the permuted label vectors are stacked into $Y \in \{0,1\}^{n \times B_2}$, and every rank sum
for every feature and every permutation is the single matrix product $R^\top Y$, which is converted
to an area under the curve using the constant positive count. The permutation $p$-value uses the
correction that keeps it away from zero \citep{phipson2010permutation},
\[
  p_j \;=\; \frac{1 + \#\{ b : T_j^{(b)} \ge T_j \}}{B_2 + 1},
\]
so the smallest reachable value is $1/(B_2+1) \approx 10^{-5}$. $B_2$ is raised if that floor is not
comfortably below the smallest targeted false discovery threshold.

Three corrections are reported side by side, because they answer different questions and a reader
should be able to see how much of a result depends on which is chosen.

\begin{itemize}[leftmargin=1.2em,itemsep=0.25em,topsep=0.35em]
\item \textbf{Benjamini and Hochberg} \citep{benjamini1995controlling}: controls the expected share of
      false calls among the calls, under independence or positive dependence.
\item \textbf{Benjamini and Yekutieli} \citep{benjamini2001control}: the same target under arbitrary
      dependence between features, at the cost of a factor $\sum_{i=1}^{p} 1/i$. Feature columns here
      are strongly dependent by construction, so this is the conservative choice here.
\item \textbf{Westfall and Young min-$P$} \citep{westfall1993resampling}: controls the chance of even
      one false call, and takes the dependence structure from the permutations themselves
      instead of bounding it. Each feature's row of the null matrix is converted to per-permutation $p$-values,
      the minimum over features is taken within each permutation, and $p_j$ is compared against that
      distribution.
\end{itemize}

Alongside them the panel reports a direct plug-in estimate of the false discovery proportion at each
raw cut-off $\alpha$ \citep{storey2003statistical},
\[
  \widehat{\mathrm{FDP}}(\alpha) \;=\;
  \frac{ B_2^{-1} \sum_{b} \#\{ j : p_j^{(b)} \le \alpha \} }
       { \max\big(1,\ \#\{ j : p_j \le \alpha \}\big) },
\]
which is the number of calls a null run makes at that cut-off divided by the number the real run
makes. It is the form of the answer a reader without a background in multiple testing can act on: at
this cut-off, this many calls, of which about this many would have appeared with no signal present.

We use $T_j$ here and never a model's own importance in its place. Permuting the labels calibrates a
multivariate importance only under the global null, so per-feature $p$-values derived from
importances lose their validity as soon as any real signal is present.

\subsection*{Tier 3: does the model need the feature}

For each $k \in R$, the model is refitted on $R \setminus \{k\}$ using the hyper-parameters,
the rows and the procedure that produced the reported model. On the evaluation rows, write $\ell_i$
for the per-row log loss and
\[
  d_i \;=\; \ell_i\big(\hat f_{-k}\big) - \ell_i\big(\hat f\big),
\]
the amount the reduced model costs on row $i$. The null hypothesis is $\mathbb{E}[d] \le 0$, tested
one-sided and paired, with a sign-flip permutation test as a check
\citep{lei2018distribution}. The panel reports the mean difference with a 95\% interval, the change in
area under the curve, and the $p$-value, corrected across the $K$ tests by Holm's step-down procedure
\citep{holm1979simple}. Where a cluster contains two or more reported features, the whole cluster is
also removed as a block and tested the same way.

This tier needs care, and every report says so. A feature strongly correlated with
another feature in the same model will cost almost nothing when removed, whatever its association
with the outcome, because what it carried is still present. The correct label for that is
substitutable rather than uninformative, and the cluster membership says what is substituting for it.
The tier is therefore most useful as an ordering, and it is decisive only where a model with ten
features loses measurably by dropping one.

\subsection*{Calibration checks}

Two runs accompany every profile.

\textbf{Global null.} The labels are replaced by a random permutation, on the training and validation
rows for Tier 1 and on the evaluation rows for Tier 2, and the tiers are re-run. The expected outcome
is that nothing is called at any correction level, that the permutation $p$-values are close to
uniform, and that selection frequencies are low and diffuse. A procedure that returns $K$ features
will still return $K$ under the null, so the informative quantity in Tier 1 is the agreement
between half-samples rather than the size of the selected set.

\textbf{Planted signal.} Labels are simulated from a logistic model on five real feature columns with
moderate coefficients, and the tiers are re-run. The expected outcome is that the five are recovered
at the top of the Tier 1 ranking and are called by Tier 2 after correction for the number of
candidates. This checks
recovery rather than error control, since neighbours of the planted five are legitimately associated
with the simulated labels. Its purpose is to give a negative result its meaning: an empty tier is
worth reporting when a planted effect of comparable size would have been found, and is worth much less
when it would not.

\subsection*{The limits of the panel}

The panel describes and does not select. It leaves the model Stage 3A produced exactly as it was, and
acting on what it says would require fresh hold-out data, since the hold-out set has by then
been used. Tier 2
detects monotone marginal association and is blind to a feature that matters only in combination with
another. Tier 1 measures the stability of a selection procedure on resamples of one dataset, which is
a weaker statement than stability across datasets. And no tier can detect a confound that predicts the
label, because such a confound survives every resampling and every permutation in the same way a real
signal does. That question belongs to the design of the study, and is settled at the first
checkpoint.

\section{Comparison with related works}
\label{app:comparison}

We provide a comparison of SeqMaestro with other approaches in Table~\ref{tab:related_work}.

\begin{table*}[t]
\centering
\caption{
Comparison of SeqMaestro with major classes of methods for interpretable
nucleotide-sequence analysis. \cmark{} indicates that a capability is supported,
\xmark{} that it is not, and N/A that the criterion is not applicable. $^*$ denotes that it depends on the specific model class.
}
\label{tab:related_work}

\resizebox{\textwidth}{!}{
\begin{tabular}{lccccc}
\toprule
Approach &
Human-readable &
Open-ended &
Intrinsic &
Interpretable functional &
Feature space \\
&
hypotheses &
hypotheses &
interpretability &
relationships &
\\
\midrule

Motif discovery
& \cmark
& \cmark
& N/A
& \xmark
& Sequence motifs \\

$k$-mer predictive models
& \cmark
& \xmark
& \cmark/\xmark$^*$
& \cmark/\xmark$^*$
& $k$-mer presence / frequency \\

Deep sequence models
& \xmark
& \xmark
& \xmark
& \xmark
& Latent representations \\

Deep learning + post-hoc interpretation
& \cmark
& \cmark
& \xmark
& \xmark
& Attributions / extracted motifs \\

Concept-based interpretation
& \cmark
& \xmark
& \xmark
& \xmark
& Predefined concepts \\

\midrule

\textbf{SeqMaestro}
& \textbf{\cmark}
& \textbf{\cmark}
& \textbf{\cmark}
& \textbf{\cmark}
& \textbf{Open-ended interpretable features} \\

\bottomrule
\end{tabular}
}

\end{table*}

\section{Running SeqMaestro under an agent}
\label{app:agent}

An agent with access to a shell and a filesystem is given three things.

\begin{enumerate}[leftmargin=1.6em,itemsep=0.3em,topsep=0.4em]
\item \textbf{A library.} Everything the Methods describes is a Python package with a test suite. The
      agent imports it rather than writing its own version, and the checkpoints verify that it did.
      If a second piece of code recomputes a feature column, there are two answers where there should
      be one, and nothing to say which of them is right.
\item \textbf{Instructions, one file per stage.} Each file states the purpose of the stage, the parts
      of the library that already do the work, the decisions the agent has to make and write down,
      what the stage must leave behind, and the conditions its checkpoint will verify.
      Table~\ref{tab:system}c lists the files, and Listing~\ref{lst:skill} shows the opening of the
      one for Stage 1.
\item \textbf{A folder for the run.} The agent may create anything inside it and nothing outside it.
      Every decision it takes is appended to a log recording the stage, the alternatives it
      considered and what the decision affects.
\end{enumerate}

What it is denied is the hold-out set. Until the models of Stage 3A and the stability tier are
fixed, the folder it can read contains none of those sequences, none of their labels and no score
computed on them.

\noindent\begin{minipage}{\linewidth}
\begin{lstlisting}[caption={Opening of the Stage 1 instructions, abridged.},label={lst:skill}]
---
name: task-construction
stage: 1
checkpoint: checkpoints/task_construction.py
---
# Stage 1. Turn the delivered data into a prediction task

## What this stage must produce
One table of (sequence, label, group) rows, a written description of the
task, and a split whose hold-out part is sealed.

## Use these rather than writing your own
seqmaestro.qc.orientation   strand convention and genome build, checked
                            against a reference
seqmaestro.grouping         a group-aware split
seqmaestro.harness.power    what this split can resolve, before any fit

## Decide, and write down why
1. The window, and the landmark it is cut from. Say what a wider window
   would cost in rows, class by class.
2. The grouping variable. Say what makes two rows dependent here.
3. What the models are told, and what is kept from them.

## Leave behind
README.md  manifest.json  splits/  description.json  open_questions.md

## The checkpoint will not pass if
- rows were discarded with no entry in the manifest
- a group appears on more than one side of the split
- description.json states a fact listed as withheld
\end{lstlisting}
\end{minipage}

\section{The prediction tasks used in this study}
\label{app:tasks}

Four biological questions, six windows. The eIF4E transcripts were windowed three ways so that the
same battery could measure how much of the answer each region gives away before a model is fitted.
Every number below comes from the record the corresponding run wrote.

\vspace{0.7em}
{\small\setlength{\tabcolsep}{4pt}
\noindent\begin{tabular}{@{}L{4.4cm}L{4.9cm}L{2.8cm}L{2.2cm}L{1.7cm}@{}}
\toprule
\textbf{Window} & \textbf{Sequences, and what the label measures} & \textbf{Training / validation / hold-out} & \textbf{Positive class, and its share} & \textbf{Split kept together on} \\
\midrule
\multicolumn{5}{@{}l}{\textbf{Polycomb nucleation}} \\[0.15em]
200 nt from the centre of a domain segment & Mouse embryonic stem cells, mm10. H3K27me3 recovery after replication, from a ChIP time course & 4,053 / 578 / 1,158 & cluster 7, 32\% & Polycomb domain \\
\addlinespace[0.35em]
\multicolumn{5}{@{}l}{\textbf{RNA polymerase II pausing}} \\[0.15em]
101 nt, read in the direction of transcription, the site at position 50 & Human, GRCh38. mNET-seq & 349,769 / 49,968 / 99,935 & paused, 50\% & none; a stratified split by site \\
\addlinespace[0.35em]
\multicolumn{5}{@{}l}{\textbf{eIF4E dependence}} \\[0.15em]
Last 30 nt of the 5$'$ untranslated region & Human transcripts. Translation measured under reduced eIF4E activity & 1,498 / 214 / 427 & factor-dependent, 48\% & gene, and identical window \\
First 150 nt of the coding region &  & 1,671 / 239 / 476 & factor-dependent, 50\% & gene, and identical window \\
Last 100 nt of the 3$'$ untranslated region &  & 1,589 / 227 / 452 & factor-dependent, 50\% & gene, and identical window \\
\addlinespace[0.35em]
\multicolumn{5}{@{}l}{\textbf{PTBP1 exon regulation}} \\[0.15em]
100 nt of intron on each side of a cassette exon, joined, the exon removed & Human, GRCh38, K562 cells. PTBP1 and PTBP2 knockdown, exon inclusion & 618 / 87 / 178 & silenced, 47\% & gene, and identical window \\
\bottomrule
\end{tabular}}

{\captionof{table}{\textbf{What each dataset is.} A window is the stretch of sequence every model sees, cut at a fixed distance from a landmark. The last column names the variable whose members were kept on one side of the split, so that no model is scored on a sequence whose relative it was fitted on.}
\label{tab:tasks}}

\vspace{1.4em}

{\small\setlength{\tabcolsep}{4pt}
\noindent\begin{tabular}{@{}L{4.0cm}L{3.5cm}L{3.2cm}L{3.2cm}cc@{}}
\toprule
\textbf{Window} & \textbf{Strongest baseline of base composition} & \textbf{Best of nine, all features} & \textbf{Best of nine, ten features} & \textbf{Ref.} & \textbf{Floor} \\
\midrule
\multicolumn{6}{@{}l}{\textbf{Polycomb nucleation}} \\[0.15em]
200 nt from the centre of a domain segment & Sixteen dinucleotide counts, 0.867 (within ten numbers: four base fractions, 0.822) & Additive, programmatic, iterative, 0.869 & Additive, programmatic, one-shot, 0.867 & 0.858 & 0.536 \\
\addlinespace[0.35em]
\multicolumn{6}{@{}l}{\textbf{RNA polymerase II pausing}} \\[0.15em]
101 nt, read in the direction of transcription, the site at position 50 & Two indicators at the site, 0.875 & Sparse linear, positional + $k$-mer, 0.978 & Additive, programmatic, iterative, 0.960 & 0.981 & 0.504 \\
\addlinespace[0.35em]
\multicolumn{6}{@{}l}{\textbf{eIF4E dependence}} \\[0.15em]
Last 30 nt of the 5$'$ untranslated region & Sixteen dinucleotide counts, 0.648 (within ten numbers: four base fractions, 0.607) & Additive, programmatic, iterative, 0.671 & Additive, programmatic, one-shot, 0.688 & 0.640 & 0.555 \\
First 150 nt of the coding region & Sixteen dinucleotide counts, 0.876 (within ten numbers: four base fractions, 0.817) & Additive, programmatic, iterative, 0.877 & Additive, programmatic, iterative, 0.872 & 0.868 & 0.552 \\
Last 100 nt of the 3$'$ untranslated region & Four base fractions, 0.754 & Additive, programmatic, iterative, 0.745 & Additive, programmatic, iterative, 0.745 & 0.726 & 0.553 \\
\addlinespace[0.35em]
\multicolumn{6}{@{}l}{\textbf{PTBP1 exon regulation}} \\[0.15em]
100 nt of intron on each side of a cassette exon, joined, the exon removed & Pyrimidine fraction, 0.743 & Additive, programmatic, iterative, 0.864 & Additive, programmatic, one-shot, 0.875 & 0.857 & 0.584 \\
\bottomrule
\end{tabular}}

{\captionof{table}{\textbf{What each window supports.} Every score is AUROC on the hold-out set. The baseline of base composition is the strongest of the Stage 1 baselines, fitted with nothing beyond a logistic
regression, and where that baseline uses more numbers than a ten-feature model is allowed, the strongest one within that allowance is given as well. The two model columns cover the nine configurations whose models can be interpreted as statements about features. \textbf{Ref.} is the best of the four gradient-boosted configurations, present only as a point of comparison. \textbf{Floor} is the smallest separation the hold-out set can tell apart from chance.}
\label{tab:tasks-results}}

\section{Features and extraction mechanisms}
\label{app:features}

\begin{figure}[p]
\centering
\includegraphics[width=0.84\linewidth]{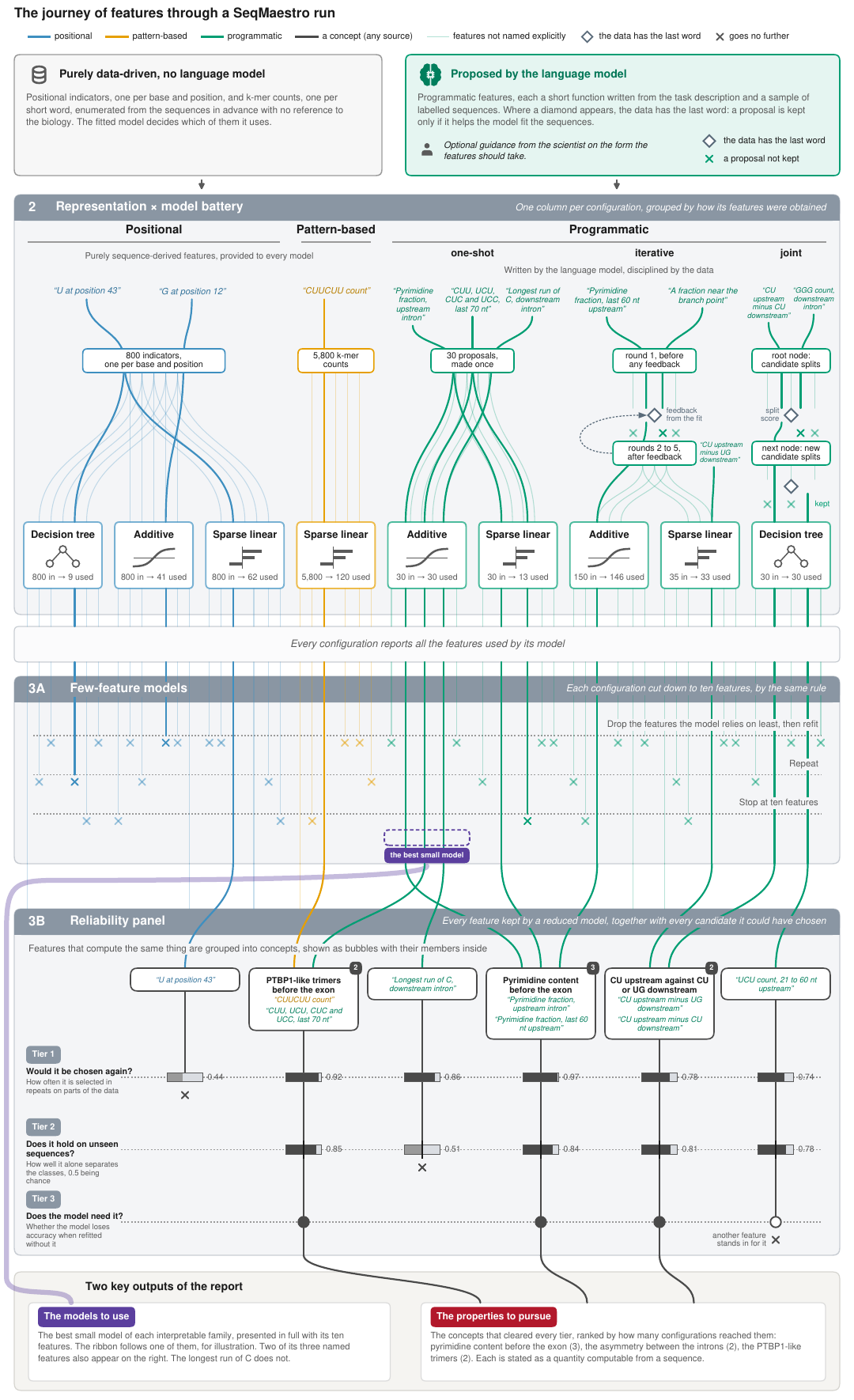}
\caption{\textbf{The journey of features through a SeqMaestro run.} A companion to
Figure~\ref{fig:method}: its stages as grey bands, and thirteen features followed through them as
coloured traces, named in the spirit of the PTBP1 task with every number illustrative. Positional
and pattern-based features are enumerated from the sequences, programmatic features are written by
a language model, and a diamond marks where the data filters the proposals, which happens at a
different point in each of the three programmatic lanes. In Stage 3A every configuration is
reduced to ten features by the same algorithm, and in Stage 3B computationally similar features
are grouped into concepts, each analysed for stability, association and necessity in turn. The best small models reach the
report with all their features, whereas only the concepts that clear every tier reach the list of
properties to pursue.}
\label{fig:flow}
\end{figure}

Figure~\ref{fig:method} presents a run stage by stage. Figure~\ref{fig:flow} complements it by
following individual features through the same stages, from the point at which they are
generated to the report. Positional and pattern-based features are derived from the sequences without any language model,
whereas programmatic features are proposed by a language model, either from the task description
alone or, in the iterative and joint settings, with the fitted models feeding back on which
proposals should be kept. The figure traces how features from each source enter the models of
Stage 2, how the reduction of Stage 3A and the reliability panel of Stage 3B narrow them down and
assess their robustness, and how the panel groups computationally similar features from different
configurations into concepts. It ends at the two outputs of the report that a feature can reach,
the best small models and the list of reliable properties. The feature names are borrowed from the
PTBP1 task but the figure is illustrative and does not correspond to a particular run.

\section{Computational requirements}
\label{app:compute}

A run costs time in two ways that behave quite differently. Fitting models on the local machine
takes longer as the sequences and the candidate features multiply. Calls to a language model depend
on how the Stage 2 battery, the grid of feature representations paired with model classes, is
configured, so a dataset several hundred times larger asks for roughly the same number of them. The six runs behind this study each ran one stage after another on a single
ten-core workstation with no GPU, and the numbers below come from the records they left behind.

\begin{figure}[t]
\centering
\includegraphics[width=0.98\linewidth]{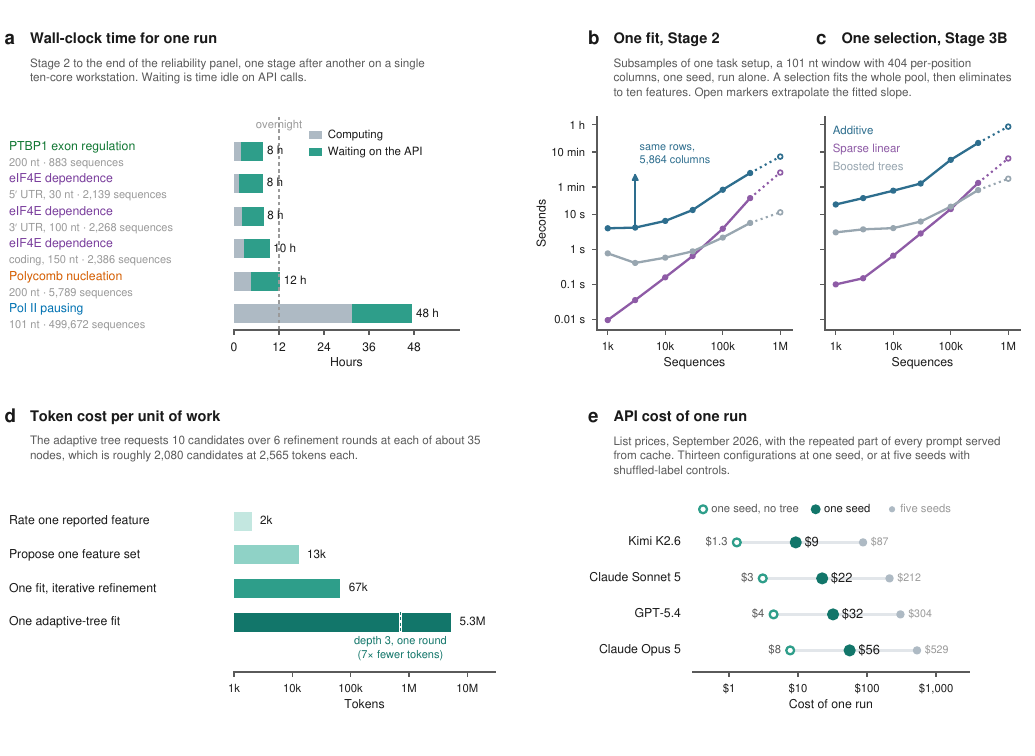}
\caption{\textbf{Computational requirements.}
\textbf{a}, Wall-clock time of each run from Stage 2 to the end of the reliability panel, split
into computation and waiting on API responses.
\textbf{b}, Time for a single fit against the number of sequences, on subsamples of one task setup,
a 101 nt window with 404 per-position columns. Open markers extrapolate the fitted slope to a
million sequences, well beyond any dataset here. The
arrow marks the same model and rows with the pool widened to 5,864 columns.
\textbf{c}, The same for one selection of the reliability panel: a fit on the whole pool, then
elimination to ten features.
\textbf{d}, Tokens consumed by each kind of request, averaged over the six runs. The dashed line on
the adaptive tree marks the same fit at depth 3 with a single round of refinement.
\textbf{e}, API cost at list prices in September 2026, with prompt caching, for the thirteen
configurations at one seed, at one seed without joint generation, and at five seeds with
shuffled-label controls.}
\label{fig:compute}
\end{figure}

\subsection*{Wall-clock time}

Five of the six task setups ran unattended and finished in eight to twelve hours. The RNA
polymerase II pausing task, with 499,672 sequences, took two days. Most of that time went on the
Stage 2 battery in every case. The reliability panel is the stage most sensitive to the size of the
dataset: under two hours on the five smaller setups, thirteen on the largest, where it ran with
seven configurations and twenty half-samples, the random halves of the data on which the panel
repeats its feature selection.

On the smaller setups, roughly three quarters of the elapsed time is spent waiting for the language
model to respond (Fig.~\ref{fig:compute}a), so the pace of a run is set by the latency of the API
rather than by the processor. The largest share of the waiting belongs to the LLM iterative feature proposer,
whose thirty fits ran one after another at five to eight minutes each. The ten fits of the adaptive
tree took over an hour each but are set up to be executed concurrently, and finished in about
eighty-five minutes together.

\subsection*{How fitting and the panel scale}

Figure~\ref{fig:compute}b keeps the sequence window and the candidate pool fixed and varies only the number
of sequences. Fitting time rises roughly in step with the sequences for the additive model, more
slowly for gradient-boosted trees, and faster for the sparse linear model, with an exponent near
1.5. Extending the fitted slopes to a million sequences, twice the largest dataset in this study,
puts a single additive fit at about seven minutes.

The reliability panel's unit of work is a selection: a fit on the whole candidate pool followed by
elimination to ten features, about twenty-five fits on a shrinking pool. Measured the same way
(Fig.~\ref{fig:compute}c), a selection costs between four and seven times a single fit and grows
with the sequences at much the same rate. With a hundred selections per configuration, as in this
study, the additive configuration on a million sequences would need about 90 hours of selection.

The number of candidate features matters as much as the number of sequences. With three thousand
sequences held fixed, an additive fit on the 404 per-position columns of a 101 nt window takes four
seconds. Adding $k$-mer counts brings the pool to 5,864 columns and the same fit to three minutes.

\subsection*{Tokens}

Taken one at a time, the requests are modest (Fig.~\ref{fig:compute}d). Rating a reported feature
takes about 2,000 tokens. A one-shot proposal takes about 13,000, which then serve all three models
fitted on it. A fit with iterative refinement takes about 67,000 across twelve calls.

Joint generation is the exception, because the tree asks for candidates at every node it expands.
Its cost is the product of three settings: the number of nodes, the rounds of refinement at each
node, and the candidates requested per round. The runs here used a depth-5 tree, five rounds and ten
candidates, which works out at roughly 2,000 candidates and 5.3 million tokens for one fit. A
depth-3 tree with a single round of refinement comes to about a seventh of that.

\subsection*{Cost}

Fitting the thirteen configurations once, at a single seed, uses about 6 million tokens. At list
prices in September 2026 that costs between \$9 and \$56 depending on the model
(Fig.~\ref{fig:compute}e). Without joint generation it is between \$1 and \$8. Five seeds, each
fitted again on shuffled labels as in this study, use about ten times as many tokens. These prices
assume prompt caching, which serves the repeated part of each prompt at a tenth of the input price.
It saves only about a fifth, because most of the bill is completions, which no cache serves.

The agent that prepares the data and drives the run adds a few tens of dollars on top, nearly all
of it for context read back from the cache at each turn.

\section{A design agnostic to the language model and the agent harness}
\label{app:harness}

\begin{figure}[!b]
\centering
\includegraphics[width=0.98\linewidth]{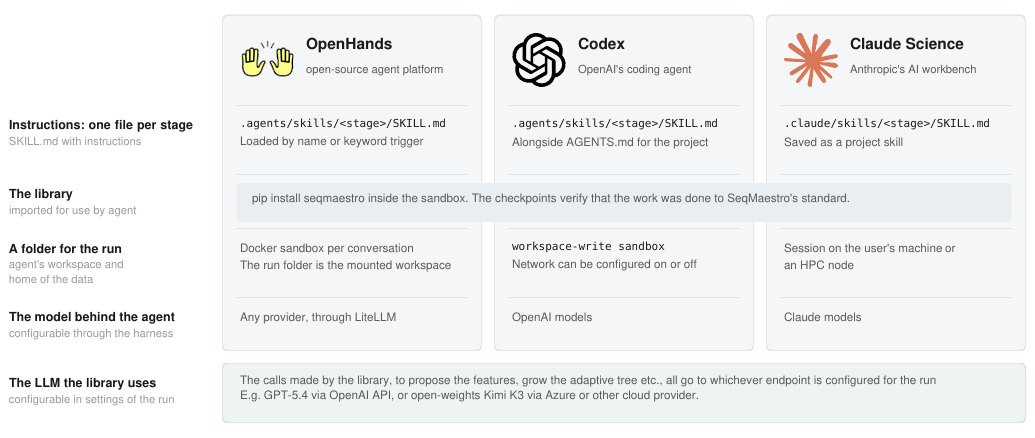}
\caption{\textbf{The method under three harnesses.} Each row is one of the things the method
requires of an agent, as listed in Appendix~\ref{app:agent}, and each column shows how one harness
provides it. The stage instructions are the same files in every column. The features named are
those documented for each product in September 2026.}
\label{fig:harness}
\end{figure}

SeqMaestro uses language models in two separate roles, and it is tied to a particular vendor in
neither. The first role is the agent that carries a run through its stages. The second is the
model that the library itself consults as it works, when it asks for programmatic features, grows
the adaptive decision tree, and rates the features that were found. This appendix explains why
neither role commits the method to a vendor, and shows how a run would be set up under three agent
environments in wide use at the time of writing: OpenHands, Codex and Claude Science
(Figure~\ref{fig:harness}).

What the method requires of the agent is set out in Appendix~\ref{app:agent}. The agent is given
the SeqMaestro library, a set of written instructions with one file per stage, and a folder in
which to work. At the end of each stage, a checkpoint verifies what the agent has left in that
folder, and the first of these checkpoints also waits for the user's approval. For the environment
that hosts the agent, these requirements reduce to three ordinary capabilities: a shell that runs
in an isolated working directory, a mechanism for loading instructions, and a means of pausing for
approval. Since every current agent environment provides all three, the choice among them is a
matter of preference and infrastructure rather than of method.

The instruction files themselves need no adaptation, because they already follow a convention
that these environments share. Each file is a Markdown document headed by a short block of
metadata, a format published as the Agent Skills specification and read by all three. OpenHands
and Codex look for such files under \texttt{.agents/skills/} within the repository, whereas Claude
looks under \texttt{.claude/skills/}, so the same files serve every environment once they have been
placed in the expected location. The metadata block that opens each file, shown in
Listing~\ref{lst:skill}, names the stage and the checkpoint script that closes it.

The two roles described above are filled independently of one another. The model that drives the
agent comes with the environment: OpenHands can reach any provider through LiteLLM, Codex uses
OpenAI's models, and Claude Science uses Claude. The calls made by the library, on the other hand,
go to whichever endpoint is named in the settings of the run, so the model that proposes and rates
features can be chosen separately from the one that drives the agent and can be replaced without
changing anything else.

\section{Understanding the interpretable models}
\label{app:reading}

This appendix is for a reader who works with sequences and does not build models. The three model
families at the centre of SeqMaestro are interpretable in the plain sense that every step from a
sequence to a prediction can be written down and checked. They differ in what they can express,
however, so the figure of one family has to be interpreted differently from the figure of another.
Each part below takes one family and answers the same three questions: how it turns a sequence into
a prediction, how to interpret the figure the report draws of it, and what it can say about the
biology that a ranked list of features cannot. The figures use the ten-feature models fitted to the
PTBP1 exon task of the Results (Fig.~\ref{fig:r4}), with one training exon followed through all
three, but nothing below is specific to that task.

\subsection*{What every model is given}

\begin{figure}[t]
\centering
\includegraphics[width=0.98\linewidth]{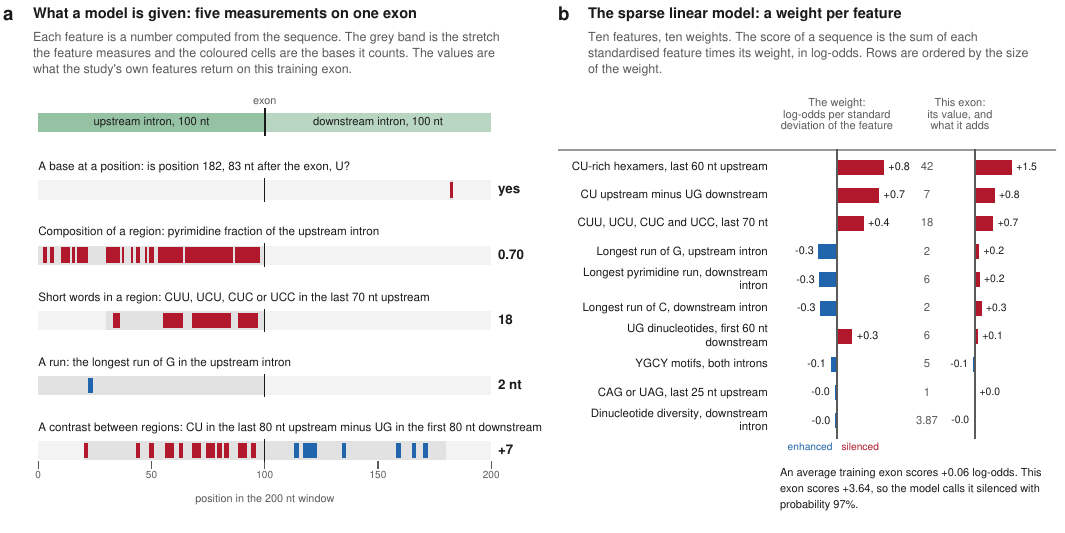}
\caption{\textbf{What a model is given, and the sparse linear model.}
\textbf{a}, Five features measured on one training exon of the PTBP1 task. The window is the two
introns flanking the exon, joined. For each feature the grey band is the stretch it measures, the
coloured cells are the bases it counts, and its value is given at the right.
\textbf{b}, The ten-feature sparse linear model of the same task. Left, the weight of each feature,
in log-odds per standard deviation of that feature. Right, the exon of panel a: its value on each
feature, and what that adds to its score relative to an average training exon. Red argues silenced
and blue enhanced, as in Fig.~\ref{fig:r4}.}
\label{fig:read-linear}
\end{figure}

No model works on the sequence itself. Each is given features, which are numbers computed from the
sequence, to learn from. Figure~\ref{fig:read-linear}a shows five features measured on one
exon. The simplest is an indicator, one or zero, for a base at a position. The other four each
summarise a stretch of the sequence in a different way: the composition of a region, the number of
short words in a region, the length of a run, and a contrast between two regions. These four are
programmatic features in the sense of the Methods, each a few lines of code written by a language
model. Most of the expressive power of the framework lies in them, since such a feature can compare
two regions, count a motif only within a set distance of a landmark, measure a run or a spacing, or
make one measurement conditional on another. Composition is the simplest case. A feature such as ``CU in
the last 80 nt upstream minus UG in the first 80 nt downstream'' already states a hypothesis about
position and asymmetry.

Table~\ref{tab:reading} sets the three families side by side. They differ in what a feature becomes
inside the model, which decides both what the model can express and how its figure is interpreted.

{\footnotesize
\setlength{\tabcolsep}{4pt}
\renewcommand{\arraystretch}{1.2}
\noindent\begin{tabular}{@{}L{2.6cm}L{4.3cm}L{4.3cm}L{4.3cm}@{}}
\toprule
 & \textbf{Sparse linear} & \textbf{Additive} & \textbf{Decision tree} \\
\midrule
A feature becomes & One weight & One curve & One question with a threshold, possibly asked more than once \\
The prediction is & The sum of weight times value & The sum of the heights looked up on the curves & The end point of a path of questions \\
The figure shows & One bar per feature & One curve per feature, over a strip of where the sequences fall & The questions, the paths, and the class balance at every point \\
It can say & Which features matter, which way, and by how much & How the effect changes across the range of a feature: a gradient, a threshold, a saturation, a reversal & Which conditions combine, and which matter only when others hold \\
It leaves to the others & Thresholds, reversals, and features acting together & Features acting together, as fitted here with one curve per feature & Gradual effects, which it renders as steps \\
\bottomrule
\end{tabular}

{\captionof{table}{\textbf{The three model families side by side.} Every model in the battery is
given the same kind of input, a set of features computed from the sequence. The families differ in
what they do with it.}
\label{tab:reading}}
}

\subsection*{The sparse linear model}

The score of a sequence is the sum of its features, each multiplied by a weight
(Fig.~\ref{fig:read-linear}b). Because the features are standardised before the fit, every weight is
in the same units, the change in score for a change of one standard deviation in the feature. The
score is in log-odds: zero is even odds between the two classes, about $+0.7$ doubles the odds of the
positive class, $+1.4$ quadruples them, and the same numbers negative favour the other class in the
same proportion. The $L_1$ penalty holds most weights at zero, so the fitted model uses a handful of
features and reports the rest as unused. That property is why the family is in the battery.

\begin{itemize}[leftmargin=1.2em,itemsep=0.2em,topsep=0.3em]
\item \textbf{Each row is a feature and the bar is its weight.} The colour is the class the feature
      argues for and the length is how much of the score the feature can account for. A weight
      argues the same way across the whole range of the feature.
\item \textbf{The right-hand column assembles the score of one sequence.} Beside each feature's
      value is what it adds relative to an average training sequence. These parts sum to the score,
      which the model converts to a probability.
\end{itemize}

The model says which features matter, in which direction and by how much, in one number per
feature. A threshold, a reversal, or two features acting together are what the next two families
add.

\subsection*{The additive model}

\begin{figure}[t]
\centering
\includegraphics[width=0.98\linewidth]{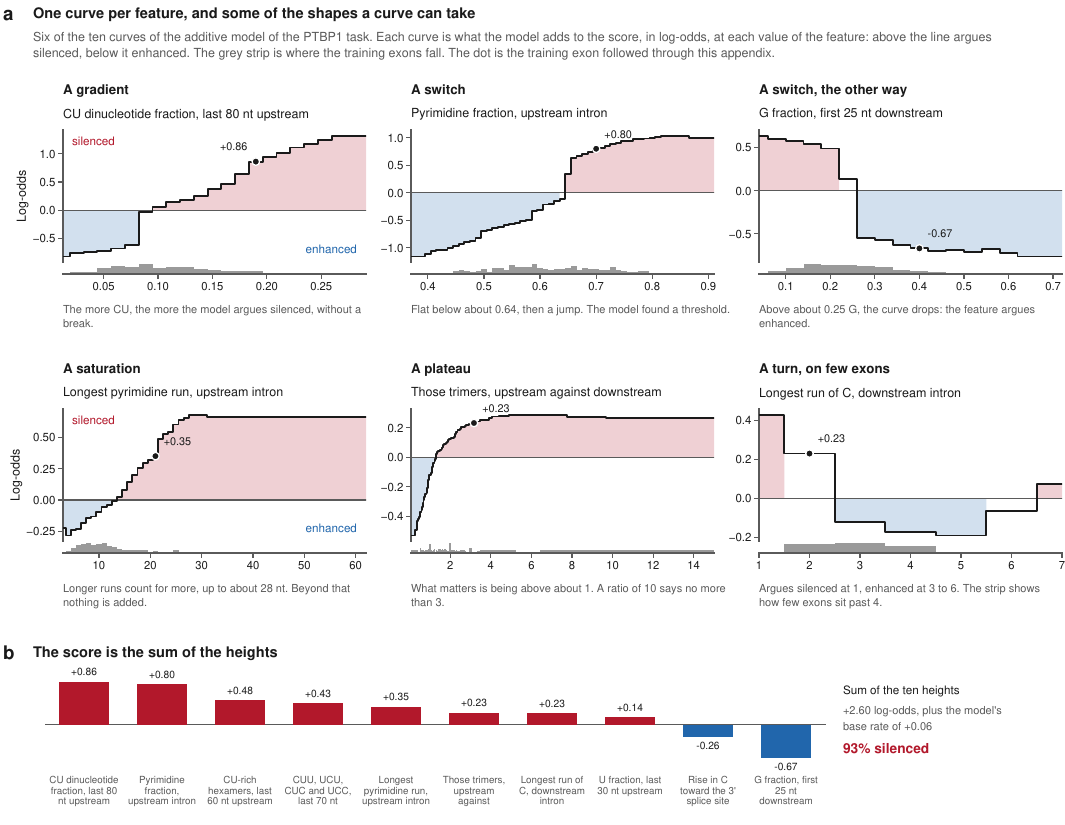}
\caption{\textbf{The additive model.}
\textbf{a}, Six of the ten curves of the additive model of the PTBP1 task, chosen for the variety
of shapes they show. Each curve is what the model adds to the score at each value of the feature, in log-odds. The
grey strip beneath is how many training exons take each value. The dot marks the exon of
Fig.~\ref{fig:read-linear}a.
\textbf{b}, The same exon's ten heights, which add up to its score.}
\label{fig:read-additive}
\end{figure}

The additive model, an explainable boosting machine, replaces each weight with a curve
(Fig.~\ref{fig:read-additive}). To score a sequence it looks up the value of each feature on that
feature's curve, takes the height there, and adds the heights. As fitted here, no term combines two
features, so the ten curves are the whole model and a figure of them hides nothing.

\begin{itemize}[leftmargin=1.2em,itemsep=0.2em,topsep=0.3em]
\item \textbf{The horizontal axis is the value of the feature and the curve is what the model adds
      at that value.} Above the line argues for one class, below it for the other. A curve may
      therefore argue both ways along its length.
\item \textbf{The strip underneath shows where the sequences are.} A steep or extreme part of a
      curve above an empty stretch of strip rests on very few sequences. The last panel of
      Fig.~\ref{fig:read-additive}a is an example, since the turn beyond a run of four comes from a
      handful of exons.
\end{itemize}

The shapes are where an additive model says more than ``higher is more''. A curve can take any
shape, since every step of it is fitted separately, and four kinds recur often enough to be worth
naming. A \textbf{gradient} means the effect grows steadily across the range, which is what a weight
would also have described. A
\textbf{switch}, flat and then a jump, means the model found a threshold, whose value is testable in
itself: an element built to cross it would show whether the effect is a switch or a gradient. A
\textbf{saturation} or a \textbf{plateau} means the effect grows to a point and then stops. That
point is again a number the biology can be asked about, such as a run beyond which extra length
earns nothing. A \textbf{turn} means the feature argues one way at low values and the
other way at high values, which is a different claim from either direction alone and one that no
single number per feature could carry. Whatever the shape, it is interpreted the same way, as
the height the model adds at each value.

\subsection*{The decision tree}

\begin{figure}[t]
\centering
\includegraphics[width=0.98\linewidth]{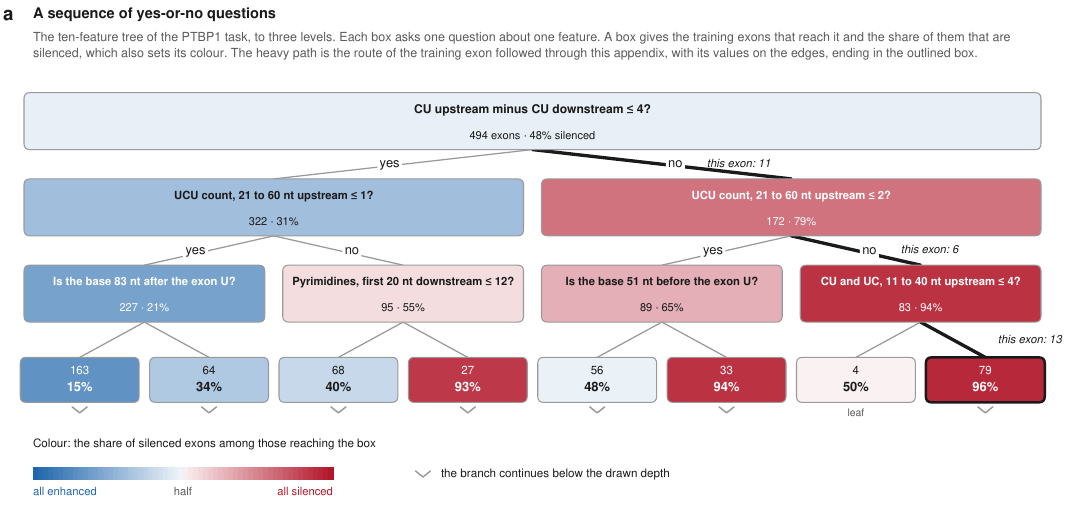}
\caption{\textbf{The decision tree.} The ten-feature tree of the PTBP1 task with programmatic features
written at each split, drawn to three levels. Each box gives the training exons that reach it and the
share of them that are silenced, which also sets its colour, red for mostly silenced and blue for
mostly enhanced. Chevrons mark branches that continue below the drawn depth. The heavy path is the
route of the exon of Fig.~\ref{fig:read-linear}a, with its values written on the edges.}
\label{fig:read-tree}
\end{figure}

A decision tree is a sequence of yes-or-no questions (Fig.~\ref{fig:read-tree}). Each box asks
whether one feature is at or below a threshold. A sequence follows one path from the top box to one
end box, taking the yes or the no branch at each question. The end box gives the prediction. The model chose each threshold from the training sequences, as the cut that best
separated the two classes among the sequences reaching that box, so a threshold is to be treated as
a hypothesis about a biological cut-off.

\begin{itemize}[leftmargin=1.2em,itemsep=0.2em,topsep=0.3em]
\item \textbf{Each box names the feature and the threshold it tests}, together with how many
      training sequences reached it and what share of them belong to the positive class. Colour
      follows that share, so a branch that separates the classes deepens in colour as it descends,
      whereas a box near fifty per cent, drawn almost white, has separated nothing yet.
\item \textbf{Chevrons mark branches that continue} below the drawn depth, so nothing is hidden about
      how much of the tree is off the page.
\item \textbf{One sequence follows one path}, which can be written out as a rule.
\end{itemize}

A path is a rule made of several conditions. This is what a tree expresses most directly: a
feature that matters only once other conditions hold. Table~\ref{tab:rules} writes
out four paths of the tree in Fig.~\ref{fig:read-tree}. Its second and third rows are the clearest
case. A single base 51 nt before the exon separates 48\% silenced from 94\%, but only among
exons that already carry more CU upstream than downstream and few UCU in the middle of the upstream
intron. On its own, that base separates the classes hardly at all. When the same feature is tested at
more than one threshold on different branches, as the count of CU and UC in the upstream intron is
in the full tree, the model is describing a graded effect in steps, or a threshold that shifts with
context.

{\footnotesize
\setlength{\tabcolsep}{4pt}
\renewcommand{\arraystretch}{1.2}
\noindent\begin{tabular}{@{}L{10.6cm}L{2.2cm}L{2.4cm}@{}}
\toprule
\textbf{The conditions along the path} & \textbf{Training exons} & \textbf{Silenced} \\
\midrule
More than 4 more CU upstream than downstream, and more than 2 UCU between 21 and 60 nt before the exon & 83 & 94\% \\
More than 4 more CU upstream than downstream, at most 2 UCU there, and U at 51 nt before the exon & 33 & 94\% \\
\quad The same conditions, with any other base at that position & 56 & 48\% \\
At most 4 more CU upstream than downstream, more than 1 UCU there, and more than 12 pyrimidines in the first 20 nt after the exon & 27 & 93\% \\
At most 4 more CU upstream than downstream, at most 1 UCU there, and a base other than U 83 nt after the exon & 163 & 15\% \\
\bottomrule
\end{tabular}

{\captionof{table}{\textbf{Four rules written out from the tree of Fig.~\ref{fig:read-tree}.} Each
row is one path from the top box to a box at the third level, given as the conditions a sequence
must meet to reach it. The second and third rows show a base that matters only in one context.}
\label{tab:rules}}
}

Because each question is chosen among the sequences that reach it, a drawn tree is one of several
that would fit the same sequences about equally well. The first tier of the reliability panel
measures this directly, by asking of every feature whether the selection would return it on another
half of the data (Methods, Stage 3B).

\section{The full analysis reports}
\label{app:reports}

Each run produces a self-contained document covering all of its stages: the preparation decisions and
what each one cost, every configuration in the battery with its shuffled-label control, the
few-feature models drawn in full with the table of features behind each figure, the whole reliability
panel including the calibration checks, and the per-feature evidence for every profiled configuration.
Measured fact and interpretation are marked apart throughout. The six documents are available at the
links below.

\begin{itemize}[leftmargin=1.2em,itemsep=0.35em,topsep=0.4em]
\item \textbf{RNA polymerase II pausing.}\\
      \url{https://www.dropbox.com/scl/fi/7v3qvhepir5n2tdhrhnkj/REPORT_polymerase_full.pdf?rlkey=ufobbwan6ojekx7kt33fv4txt&dl=0}
\item \textbf{Polycomb nucleation.}\\
      \url{https://www.dropbox.com/scl/fi/zcssccvbsmcc7ipc2nr6m/REPORT_nucleation_full.pdf?rlkey=dsxu0sex6rctfp8nejdgs4vnx&dl=0}
\item \textbf{eIF4E dependence, first fifty codons of the coding region.}\\
      \url{https://www.dropbox.com/scl/fi/eremnynkpsaojpy1i5e14/REPORT_mrna_factor_cds_full.pdf?rlkey=udvz1e5s7j88p4xu9nw9tlyya&dl=0}
\item \textbf{eIF4E dependence, the last thirty bases of the 5$'$ untranslated region.}\\
      \url{https://www.dropbox.com/scl/fi/pgjc6t4n0okbhj0c46xd7/REPORT_mrna_factor_utr5_full.pdf?rlkey=2rh05lnpfo0wmz61n7gnmya8y&dl=0}
\item \textbf{eIF4E dependence, the last hundred bases of the 3$'$ untranslated region.}\\
      \url{https://www.dropbox.com/scl/fi/vllmqz0vabvtgw02elwh1/REPORT_mrna_factor_utr3_full.pdf?rlkey=nguq2terr7dzg28vucv79dh8o&dl=0}
\item \textbf{PTBP1 exon regulation.}\\
      \url{https://www.dropbox.com/scl/fi/ar1hjl96umy4kmjjpz8s0/REPORT_ptbp1_exon_regulation_full.pdf?rlkey=445u76ji0i28cl2u85bb4yoou&dl=0}
\end{itemize}


\end{document}